\documentclass[iicol, sn-basic]{sn-jnl}

\usepackage{bm}
\usepackage{booktabs}
\usepackage{amsfonts}
\usepackage{amsmath}
\usepackage{MnSymbol}
\usepackage{amssymb}
\usepackage{amsthm}
\usepackage{bbm}
\usepackage{color}
\usepackage{graphicx}
\usepackage{subfigure}
\usepackage{bbding}
\usepackage{url}

\newtheorem{Theorem}{Theorem}  

\newtheorem{Assumption}{Assumption}

\usepackage{multirow}

\newcommand{\I}{{-1}}
\def\+#1{\mathbb{#1}}
\def\*#1{\mathbf{#1}}

\jyear{2021}%

\theoremstyle{thmstyleone}%
\theoremstyle{thmstyletwo}%

\theoremstyle{thmstylethree}%

\begin{document}

\title[Article Title]{Domain-Specific Bias Filtering for Single Labeled Domain Generalization}

\author[1]{\fnm{Junkun} \sur{Yuan}}\email{yuanjk@zju.edu.cn}
\equalcont{These authors contributed equally to this work.}
\author[2]{\fnm{Xu} \sur{Ma}}\email{maxu@zju.edu.cn}
\equalcont{These authors contributed equally to this work.}
\author[3]{\fnm{Defang} \sur{Chen}}\email{defchern@zju.edu.cn}
\author*[4]{\fnm{Kun} \sur{Kuang}}\email{kunkuang@zju.edu.cn}
\author[5]{\fnm{Fei} \sur{Wu}}\email{wufei@zju.edu.cn}
\author[6]{\fnm{Lanfen} \sur{Lin}}\email{llf@zju.edu.cn}

\affil*[1]{\orgdiv{College of Computer Science and Technology}, \orgname{Zhejiang University}, \orgaddress{\city{Hangzhou}, \country{China}}}


\abstract{Conventional Domain Generalization (CDG) utilizes multiple labeled source datasets to train a generalizable model for unseen target domains. However, due to expensive annotation costs, the requirements of labeling all the source data are hard to be met in real-world applications. In this paper, we investigate a Single Labeled Domain Generalization (SLDG) task with only one source domain being labeled, which is more practical and challenging than the CDG task. A major obstacle in the SLDG task is the discriminability-generalization bias: the discriminative information in the labeled source dataset may contain domain-specific bias, constraining the generalization of the trained model. To tackle this challenging task, we propose a novel framework called Domain-Specific Bias Filtering (DSBF), which initializes a discriminative model with the labeled source data and then filters out its domain-specific bias with the unlabeled source data for generalization improvement. We divide the filtering process into (1) feature extractor debiasing via k-means clustering-based semantic feature re-extraction and (2) classifier rectification through attention-guided semantic feature projection. DSBF unifies the exploration of the labeled and the unlabeled source data to enhance the discriminability and generalization of the trained model, resulting in a highly generalizable model. We further provide theoretical analysis to verify the proposed domain-specific bias filtering process. Extensive experiments on multiple datasets show the superior performance of DSBF in tackling both the challenging SLDG task and the CDG task.}




\maketitle

\begin{figure*}[t]
    \centering
    \includegraphics[trim={0cm 0cm 0cm 0cm},clip,width=2.\columnwidth]{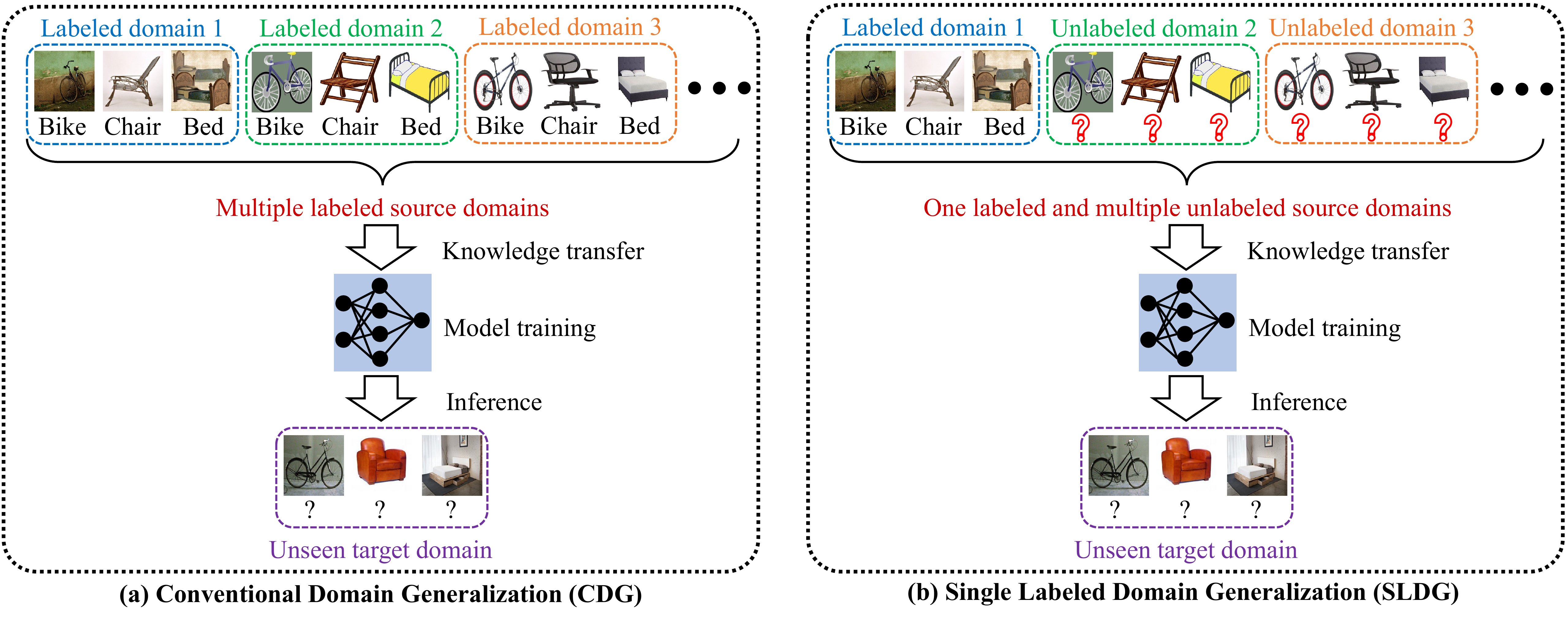} 
    \caption{Comparison between the CDG (a) and the introduced SLDG tasks (b) for visual recognition. The latter is more practical for dealing with the problem of high annotation costs in real-world applications, yet challenging, because only one of the multiple source datasets is labeled, which may lead to a serious problem of discriminability-generalization bias.}
    \label{fig-sldg}
\end{figure*}

\section{Introduction}\label{int}
Deep learning based \emph{supervised learning} (SL) and \emph{semi-supervised learning} (SSL) have made great progress in recent years  \citep{lecun2015deep, yang2021survey}. However, their success heavily relies on the independent and identically distributed (i.i.d.) assumption \citep{vapnik1992principles}, while the training (source) and test (target) datasets are usually sampled from different distributions in real-world applications, which is known as \emph{dataset shift} \citep{quionero2009dataset}. 
To address this problem, \emph{domain adaptation} (DA) \citep{ben2010theory} and \emph{domain generalization} (DG) \citep{blanchard2011generalizing} are formulated and many effective methods \citep{wang2021domain, lin2020multi, xu2021neutral, wang2020attention, ding2017deep} are proposed to improve the out-of-domain generalization ability of the model.

Typical research fields of DA, such as \emph{unsupervised domain adaptation} (UDA) \citep{xu2021neutral, zhang2020collaborative, li2020aligning, li2020generating, long2018conditional, Zhang2019BridgingTA}, \emph{multi-source domain adaptation} (MSDA) \citep{zuo2021attention, peng2019moment, zhang2015multi, zhao2018adversarial, wang2020attention}, and \emph{multi-target domain adaptation} (MTDA) \citep{Chen2019BlendingTargetDA, Liu2020OpenCD, wang2020attention, Gong2013ReshapingVD, Yu2018MultitargetUD, gholami2020unsupervised} suppose that both the source and the target datasets are available for model training. For each new target domain, they have to re-collect target data and use it to repeat the training process, which is expensive, time-consuming, or even infeasible. For example, an autonomous driving car can not know in advance which environment (i.e., domain), it will enter. 
DG is thus proposed to learn a generalizable model by incorporating the invariance across multiple labeled source domains without accessing any target data. However, labeling all the source data is laborious, and most of the previous DG methods \citep{ding2017deep, balaji2018metareg, Dou2019DomainGV, Wang2020LearningFE, Zhao2020DomainGV, Matsuura2020DomainGU} do not make full use of the information contained in massive unlabeled data. 
Then, a more practical problem arises: Is it possible to perform domain generalization with only one labeled source dataset as well as multiple unlabeled source datasets? 
For example, we may train a skin lesion classification model \citep{Li2020DomainGF} by using a labeled skin lesion dataset from a central hospital. Meanwhile, we would like to further improve the generalization ability of the model by employing abundant data from other local hospitals, but the additional data may be unlabeled due to expensive annotation costs.

In this paper, in addition to the Conventional Domain Generalization (CDG) task with multiple labeled source domains, we further investigate a more practical task, namely Single Labeled Domain Generalization (SLDG), where only one of the multiple source domains is labeled (see Fig. \ref{fig-sldg}). The single labeled multi-source data puts a serious obstacle in the path of generalization learning, which we call the \emph{discriminability-generalization bias}: the discriminative information in the labeled source domain may contain domain-specific bias, constraining the out-of-domain generalization ability of the trained model. Thus, how to train a discriminative model while removing its domain-specific bias for guaranteeing generalization is the key to solve this challenging task. 

To address this problem, we propose a novel framework called Domain-Specific Bias Filtering (DSBF) for the SLDG task. Specifically, it initializes a discriminative model with the labeled source data and then filters out domain-specific bias in the initialized model with the unlabeled source data for generalization improvement, corresponding to a model initialization stage and a bias filtering stage, respectively. The bias filtering stage consists of (1) feature extractor debiasing via k-means clustering-based semantic feature re-extraction and (2) classifier rectification through attention-guided semantic feature projection. Our method DSBF unifies the exploration of labeled and unlabeled source data to enhance the discriminability and generalization of the trained model, resulting in a highly generalizable model, as verified by theoretical analyses. Extensive experiments on multiple datasets consistently show the superior performance of the proposed DSBF framework for the SLDG task. Moreover, we also verify the effectiveness of it for the CDG task with multiple labeled source domains.

Our main contributions are summarized as: (1) We investigate a practical and challenging generalization task, namely Single Labeled Domain Generalization (SLDG) with only one source domain being labeled, towards the real scenarios where massive unlabeled data is available for generalizable model training. (2) We propose a novel framework called Domain-Specific Bias Filtering (DSBF) to tackle the SLDG task by unifying the exploration of the labeled and the unlabeled source data, which consists of model initialization and bias filtering that enhances the discriminability and generalization ability of the model, respectively. (3) We verify the proposed method DSBF with theoretical analyses. Extensive experiments on multiple datasets consistently show its superior performance in tackling the SLDG task. Our method can be easily extended to the CDG task and also achieves state-of-the-art performance. 

\section{Related Work}\label{sec:rel}
\subsection{Supervised and Semi-Supervised Learning}
In recent years, deep learning based supervised learning (SL) has been widely employed in a variety of applications \citep{lecun2015deep}. It considers the principle of empirical risk minimization (ERM) \citep{vapnik1992principles} that a model with low empirical risk on a labeled training dataset is supposed to generalize well on a test dataset. Due to the expensive annotation costs, lots of recent works \citep{Tarvainen2017MeanTA, SohnBCZZRCKL20, yasarla2021semi, wang2020enaet} focus on semi-supervised learning (SSL) \citep{yang2021survey} that utilizes both the labeled and the unlabeled data for model training. For example, \cite{Tarvainen2017MeanTA} train a student model with a classification cost on the labeled data, and use a consistency cost to make the outputs of the student and a teacher be consistent on the unlabeled data for effectively capturing discriminative information. Although the general SSL methods make full use of both the labeled and the unlabeled data for training discriminative models, they assume that all the datasets are sampled from the same distributions, which may make the trained models suffer from significant performance degradation on the test (target) datasets in real scenarios. In comparison, the SLDG task that we investigate aims to train a generalizable model using both the labeled and the unlabeled source data, for better generalization on unknown target domains with different statistical distributions.

\subsection{Domain Adaptation}
Unsupervised domain adaptation (UDA) \citep{ben2010theory, bellitto2021hierarchical, chen2021scale, dai2020curriculum, gong2014learning, ho2014model, hoffman2014asymmetric, huang2021unsupervised, kan2014domain, li2021unsupervised, shen2021cdtd, sindagi2017domain, xu2016hierarchical, yamada2014domain, zhao2021madan, zheng2021rectifying} is a prevailing direction to DA that addresses the dataset shift between a labeled source domain and an unlabeled target domain. Considerable progress has been made in UDA. A large proportion of them reduces divergence between the source and target domains via adversarial learning \citep{zhang2020collaborative, li2020generating, ganin2016domain, long2017deep, long2018conditional, saito2018maximum, Zhang2019BridgingTA} or directly minimizing domain discrepancy with a metric like Maximum Mean Discrepancy (MMD) \citep{li2020aligning, long2015learning, long2017deep}. 
These methods may fail to leverage the available multiple source datasets, leading to insufficient generalization learning.

Increasing works \citep{zuo2021attention, peng2019moment, zhang2015multi, zhao2018adversarial, wang2020attention} thus focus on the multi-source domain adaptation (MSDA) \citep{ben2010theory} task, where multiple labeled source datasets from different domains are provided for model adaptation. For example, some works \citep{zuo2021attention, wang2020attention} present an attention-based strategy to reduce domain divergence in the semantic feature space by using the multiple source datasets and an elaborate attention module. Multi-target domain adaptation (MTDA) is another research field of DA, which extends UDA to multiple \citep{Gong2013ReshapingVD, gholami2020unsupervised, Yu2018MultitargetUD, wang2020attention, Chen2019BlendingTargetDA, Yu2018MultitargetUD}, continuous \citep{gong2019dlow, mancini2019adagraph, wu2019ace}, and latent \citep{hoffman2012discovering, xiong2014latent, mancini2019inferring, Liu2020OpenCD} target domains. Among them, \cite{Chen2019BlendingTargetDA} introduce blending-target domain adaptation (BTDA) that aims to adapt the model to a mixed target distribution where the multi-target proportions are unobservable. \cite{Liu2020OpenCD} assume the target domain is a compound of multiple homogeneous domains without domain labels and employ model predictions as the pseudo labels of the unlabeled data to enable a curriculum learning process. 
Although annotation costs of the target dataset are avoided in the above DA researches, the requirements of re-collecting target data and training model for each new target domain still hinder their applications in real scenarios. In contrast, we aim to train a generalizable model that can directly generalize to unseen target domains in the investigated SLDG task. 
Note that despite both the MTDA task and our SLDG task assume one labeled dataset and multiple unlabeled datasets, MTDA mainly aims to improve the performance of the model on the seen unlabeled target domains (which can be used for both training and inference), but SLDG aims to improve the performance of the model on unseen target domains (which can only be used for inference).

\subsection{Domain Generalization}
Recently, domain generalization (DG) \citep{blanchard2011generalizing} attracts great interest, which learns to extract domain invariance from multiple labeled source datasets and trains a generalizable model to unseen target domains. Since the DG task is similar to meta-learning \citep{schmidhuber1987evolutionary}, some works \citep{balaji2018metareg, Dou2019DomainGV, Li2019EpisodicTF} employ a meta-learning-based strategy that trains the model on a meta-train dataset and continues to improve the model generalization on a meta-test dataset, both the datasets are constructed from the available labeled multi-source data. Meanwhile, a lot of effort has gone into data augmentation techniques \citep{Carlucci2019DomainGB, Wang2020LearningFE}. The latent idea of these works is the augmented data generates various new domains, and the models trained on these generated domains could be more generalizable. Similar to DA, some recent DG works \citep{Zhao2020DomainGV, Matsuura2020DomainGU, zhou2020learning} use adversarial learning to learn discriminative and domain-invariant semantic feature representations that can be applied to different domains. Other strategies like normalization \citep{Seo2020LearningTO, zhou2021domain} and else \citep{HuangWXH20, Yuan2022LabelEfficientDG, DBLP:conf/kdd/QianXLZJLZC022, qiao2020learning, yuan2021learning, Miao2022DomainGV, ding2017deep, yuan2021collaborative, DBLP:conf/mm/ZhangJWKZZYYW20} are also considered in the DG research fields. 
These methods may require fully labeled multi-source data, which is hard to be satisfied due to the high annotation costs. 

\cite{qiao2020learning} present to perform domain generalization with only one labeled source domain, and design a meta-learning scheme with an auto-encoder for model training. 
Besides, some data augmentation \citep{Volpi2018GeneralizingTU, Carlucci2019DomainGB, Wang2020LearningFE} and gradient-based \citep{HuangWXH20} methods may also be extended to the one-labeled-source setting. However, they fail to leverage the unlabeled data, which might be abundant in real scenarios, to further boost the out-of-distribution generalization of the model.

Therefore, in addition to the Conventional Domain Generalization (CDG) setting, we further investigate a more practical task called Single Labeled Domain Generalization (SLDG). The challenging SLDG task only assumes one source dataset to be labeled, and other unlabeled source datasets are further exploited to improve the out-of-distribution generalization of the model.

A related task is the Semi-Supervised Domain Generalization (SSDG) \citep{zhou2021semi}. Both the SSDG and our SLDG tasks aim to train a generalizable model using partially-labeled source data. However, they are different in the following aspects. 
(1) Problem definition: SSDG assumes that partial samples are labeled in each source domain; but our SLDG task considers that only one source domain is labeled while other domains are totally unlabeled. 
(2) Solution direction: based on the different definitions, \cite{zhou2021semi} perform semi-supervised training under the consideration of domain shift by extending FixMatch via uncertainty and style consistency learning; but we learn a discriminative model from the labeled dataset and then filter out bias and boost generalization using the unlabeled datasets, corresponding to the model initialization and bias filtering stages, respectively.
(3) Application scenario: SSDG focuses on the scenario that multiple partially-labeled datasets are given for generalization learning; but our SLDG task is introduced towards the scenario that a labeled dataset is given for learning a predictive yet biased model, meanwhile, multiple semantically-relevant but unlabeled datasets are available for further boosting its out-of-distribution generalization performance.

\subsection{Attention Mechanism}
Attention \citep{BahdanauCB14} is first introduced in natural language processing for deciding which parts of a sentence should be paid more attention to. It is widely applied in various fields \citep{wang2017residual, zhang2019self, fu2019dual, devlin2018bert}. 
Self-attention/intra-attention \citep{VaswaniSPUJGKP17} is a specific form of the attention mechanism, which learns a representation of a sequence by reweighting its different positions according to their importance. A general process of the self-attention consists of three steps: (1) Getting the embeddings of \emph{query}, \emph{key}, and \emph{value} from the original sequence. (2) Obtaining normalized weights by calculating the similarity between the query and the key. (3) Weighting the value. 
For example, \cite{fu2019dual} capture rich contextual dependencies in both spatial and channel dimensions by using a position attention module and a channel attention module, selectively aggregating spatial and channel features for obtaining more effective representations. 
In the inter-domain attention module of our method DSBF, we let the key-value pairs be constructed from the semantic features of one source domain and the query be constructed from the semantic features of other source domains. In this way, the similar semantic information is automatically enhanced for generalization improvement.

\section{Problem Setup}
In Conventional Domain Generalization (CDG) task, we may assume that there are $K$ labeled multi-source datasets $\{\mathcal{D}^{j}\}_{j=1}^{K}$ with $n^{j}$ samples in the $j$-th dataset, i.e., $\mathcal{D}^{j}=\{(\boldsymbol{x}_{i}^{j},y_{i}^{j})\}_{i=1}^{n^{j}}$. Any information of the target domain $\mathcal{D}^{K+1}$ is not provided during the model training process. The source datasets $\mathcal{D}^{1},...,\mathcal{D}^K$ and the target dataset $\mathcal{D}^{K+1}$ are sampled from different distributions $P(X^{1},Y^{1}),...,P(X^{K},Y^{K}),P(X^{K+1},Y^{K+1})$, respectively, which are defined on input and label joint space $\mathcal{X}\times\mathcal{Y}$. 
The goal of the CDG task is to use the fully labeled source datasets $\{\mathcal{D}^{j}\}_{j=1}^K$ to train a predictive model that can perform well on the unseen target dataset $\mathcal{D}^{K+1}$.

In this paper, we further introduce a more challenging task, i.e., Single Labeled Domain Generalization (SLDG). SLDG also aims to improve the generalization performance on the unseen target domain, but only the first source dataset $\mathcal{D}^{1}=\{(\boldsymbol{x}_{i}^{1},y_{i}^{1})\}_{i=1}^{n^{1}}$ is assumed to be labeled, the others $\{\mathcal{D}^{j}=\{\boldsymbol{x}_{i}^{j}\}_{i=1}^{n^{j}}\}_{j=2}^{K}$ are supposed to be unlabeled. 

The main challenge in the SLDG task is the discriminability-generalization bias. That is, when we use the labeled source data to train a discriminative model for object recognition, the domain-specific bias in the labeled source data would mislead the model, constraining its generalization performance on other domains. Therefore, it is vital to train a discriminative model using the labeled source data while removing its domain-specific bias for generalization improvement.

\section{Methodology}
\begin{figure*}[t]
    \centering
    \includegraphics[trim={0cm 0cm 0cm 0cm},clip,width=2\columnwidth]{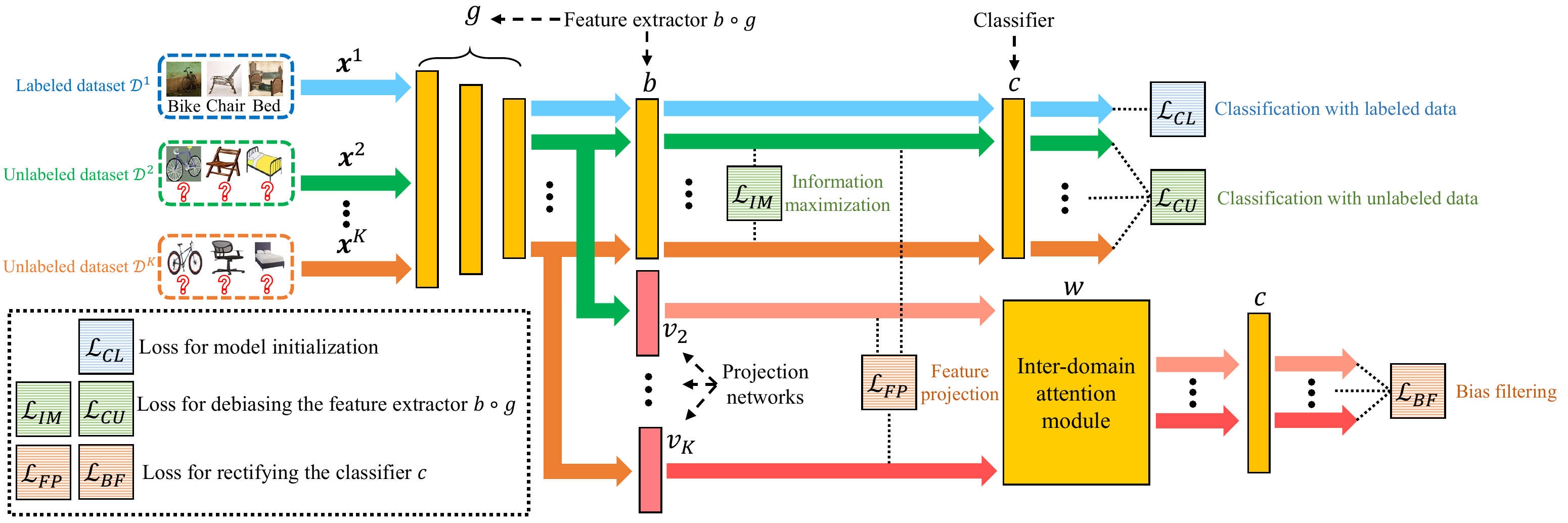} 
    \caption{Our proposed Domain-Specific Bias Filtering (DSBF) framework. The whole model consists of a feature extractor $b\circ{g}$, a classifier $c$, projection networks $\{v_{j}\}_{j=2}^{K}$, and an attention module $w$. We employ $\{v_j\}_{j=2}^{K}$ and $w$ only for classifier rectification in training. After training, we only use the trained $\hat{c}\circ{\hat{b}\circ{\hat{g}}}$ for inference on out-of-distribution target domains.}
    \label{fig-framework}
\end{figure*}

\begin{algorithm}[t]  
    \caption{Domain-Specific Bias Filtering}  
    \label{alg}
    \begin{algorithmic}[1]
        \Require 
        A labeled source dataset $\mathcal{D}^{1}$, unlabeled source datasets $\{\mathcal{D}^{j}\}_{j=2}^{K}$, a backbone $g$ and a network $b$ with parameter $\theta_{g}$ and $\theta_{b}$ of the feature extractor, a classifier $c$ with parameter $\theta_{c}$, semantic feature projection networks $\{v_{j}\}_{j=2}^{K}$ with parameters $\{\theta_{v_{j}}\}_{j=2}^{K}$, initialization/filtering iterations $M$/$N$.
    \Ensure  
        Well-trained $\hat{g}$, $\hat{b}$, and $\hat{c}$ for inference.
    \State Initialize SGD optimizers and parameters;
    \For{$iter=1$ to $M$} // model initialization
        \State Sample a batch data from $\mathcal{D}^{1}$;
        \State Update $\theta_{g}$, $\theta_{b}$, $\theta_{c}$ by minimizing Eq. (\ref{equation-cl});
    \EndFor
    \For{$iter=1$ to $N$}  // bias filtering
        \State Sample a batch data from $\mathcal{D}^{j}$, $j=1,...,K$;
        \State Get pseudo labels via Eq. (\ref{equation-ak0}-\ref{equation-ak1});
        \State Update $\theta_{g}$, $\theta_{b}$ by minimizing Eq. (\ref{equation-im}-\ref{equation-cu});
        \State Update $\{\theta_{v_{j}}\}_{j=2}^{K}$ by minimizing Eq. (\ref{equation-fp});
        \State Update $\theta_{c}$ by minimizing Eq. (\ref{equation-bf}).
    \EndFor
  \end{algorithmic}
\end{algorithm}

We address the SLDG task by proposing Domain-Specific Bias Filtering (DSBF). It initializes a discriminative model using the labeled source data and filters out domain-specific bias in the initialized model using the unlabeled source data for generalization improvement, which corresponds to a \emph{model initialization} stage and a \emph{bias filtering} stage. The bias filtering consists of (1) \emph{feature extractor debiasing} using the unlabeled data and its pseudo labels obtained via k-means clustering and (2) \emph{classifier rectification} through attention-guided semantic feature projection. Our framework and algorithm are shown in Fig. \ref{fig-framework} and Alg. \ref{alg}, respectively. We then introduce the details of the two stages of the DSBF method in the following.

\subsection{Model Initialization}
To initialize a discriminative model, we use the labeled source data $\mathcal{D}^{1}$ to pretrain the feature extractor $b\circ{g}$ and the classifier $c$ for learning to extract semantic features of the data and classifying the extracted features to the corresponding categories, respectively. 
The used cross-entropy classification loss of the labeled source data for initializing the model, i.e., $b\circ{g}$ and $c$, is
\begin{equation}\label{equation-cl}
    \mathcal{L}_{CL}=-\sum_{r=1}^{C}\boldsymbol{y}_{r}^{1}\log{f_{r}^{b}(\boldsymbol{x}^{1})},
\end{equation}
where $f^{b}:=c\circ{b}\circ{g}$ outputs softmax classification of the data, and $f^{b}_{r}$ is the $r$th dimension output of $f^{b}$. $\boldsymbol{y}_{r}^{j}$ is the $r$-th dimension of one-hot encoding of the labels $y^{j}\in\{1,...,C\}$ of domain $j$, where the correct class is ``1", otherwise is ``0". 

After model initialization, the feature extractor $b\circ{g}$ and the classifier $c$ are pretrained to extract semantic features of the data and use them for classification, respectively. However, they may be misled by the domain-specific bias of the labeled source data. Thus, we utilize the unlabeled data to filter out the domain-specific bias in the initialized model for generalization improvement. 

\subsection{Bias Filtering}
The bias filtering consists of feature extractor debiasing and classifier rectification. In feature extractor debiasing, we aim to train the feature extractor using the unlabeled source data for reducing the bias of the feature extractor towards the labeled source data, and hence obtaining a more robust model. Since the unlabeled source data does not have ground-truth labels, we exploit k-means clustering to obtain the pseudo labels and use them to train the feature extractor for effective semantic feature re-extraction. In classifier rectification, we project the semantic features of the unlabeled source data to the semantic features of the labeled source data, and then use the projection features to predict the labels of the labeled source data. Since the projection features only contain the bias of the unlabeled source data (because it is obtained by feeding the features of the unlabeled source data to the projection networks) while the labels only contain the bias of the labeled source data, optimizing the classifier with supervised loss can debias it, which is verified by the theoretical analyses in Section \ref{sectio-theoretical-insights}. We introduce an inter-domain attention module to further capture the similarities among domains and boost generalization performance.

\subsubsection{Feature Extractor Debiasing}
We obtain pseudo labels of the unlabeled data to facilitate the following feature extractor debiasing and classifier rectification processes. Inspired by recent works \citep{Kang2019ContrastiveAN, Liang2020DoWR} on deep clustering \citep{caron2018deep}, we adopt k-means clustering to assign pseudo labels $\{\hat{\boldsymbol{y}}^{j}\}_{j=2}^{K}$ for the unlabeled source datasets $\{\boldsymbol{x}^{j}\}_{j=2}^{K}$. Specifically, we first get a centroid $\boldsymbol{a}_{(j,r)}^{(0)}$ of each class $r$ for the semantic features of each unlabeled domain $j$ by softly assigning each sample $\boldsymbol{x}^{j}$ to it with model prediction-based score $f_r^b(\boldsymbol{x}^{j})$, that is
\begin{equation}\label{equation-ak0}
        \boldsymbol{a}_{(j,r)}^{(0)}=\frac{\sum_{\boldsymbol{x}^{j}}{f_{r}^{b}(\boldsymbol{x}^{j})}b\circ g(\boldsymbol{x}^{j})}{\sum_{\boldsymbol{x}^{j}}{f_{r}^{b}(\boldsymbol{x}^{j})}}.
\end{equation}
The centroid $\boldsymbol{a}_{(j,r)}^{(0)}$ represents the semantic feature distribution of each class $r$ in domain $j$, which is used to assign the initial pseudo label $d^{j}$ for the samples $\boldsymbol{x}^{j}$, that is
\begin{equation}\label{equation-d}
    d^{j}=\arg\min_{r}{\rm{dist}}(b\circ g(\boldsymbol{x}^{j}),\boldsymbol{a}_{(j,r)}^{(0)}),
\end{equation}
where ${\rm{dist}}(\cdot,\cdot)$ measures the cosine distance. Similarly, we then get updated centroid $\boldsymbol{a}_{(j,r)}^{(1)}$ and final pseudo labels $\hat{y}^{j}$ by
\begin{equation}\label{equation-ak1}
    \begin{aligned}
    \boldsymbol{a}_{(j,r)}^{(1)}&=\frac{\sum_{\boldsymbol{x}^{j}}{1}(d^{j}=r)b\circ g(\boldsymbol{x}^{j})}{\sum_{\boldsymbol{x}^{j}}{{1}(d^{j}=r)}},\\
    \hat{y}^{j}&=\arg\min_{r}{\rm{dist}}(b\circ g(\boldsymbol{x}^{j}),\boldsymbol{a}_{(j,r)}^{(1)}).
    \end{aligned}
\end{equation}
The pseudo labels $\hat{y}^{j}$ can be transformed into one-hot encoding $\hat{\boldsymbol{y}}^{j}$.
We consider the ideal form of the softmax outputs of $c$ should be like one-hot encoding for each sample, and be distinct for samples from different classes.
Therefore, we improve the clustering performance by optimizing $g$, $b$ with information maximization constraint \citep{kundu2020universal, Liang2020DoWR} loss:
\begin{equation}\label{equation-im}
    \begin{aligned}
        \mathcal{L}_{IM}=&\frac{1}{K-1}\sum_{j=2}^{K}\left\{\underbrace{\sum_{r=1}^{C}t_{r}\log{t_{r}}}_{\text{diverse class}}\right.
        \underbrace{\left.-\mathbb{E}[\sum_{r=1}^{C}u_{r}\log{u_{r}}]\right\}}_{\text{concentrated sample}},
    \end{aligned}
\end{equation}
where $t_{r}=\mathbb{E}[f_{r}^{b}(\boldsymbol{x}^{j})]$ and $u_{r}=f_{r}^{b}(\boldsymbol{x}^{j})$. The first term on the r.h.s. of Eq. (\ref{equation-im}), i.e., negative expected entropy of population, makes the outputs of $c$ diverse at the class level. The second term on the r.h.s. of Eq. (\ref{equation-im}), i.e., expected entropy of individual, makes the outputs of $c$ be concentrated at the sample level. Through minimizing $\mathcal{L}_{IM}$, we encourage the unlabeled samples with closer distance group together, meanwhile, the samples far away are further separated. It improves the clustering performance and allows us to obtain more accurate pseudo labels for bias filtering. 

After clustering, we obtain the pseudo labels of the unlabeled source data. To debias the feature extractor $b\circ{g}$, we present to retrain it with the average classification loss of all the unlabeled source datasets, thus re-extract the semantic feature of the source data, that is,
\begin{equation}\label{equation-cu}
    \mathcal{L}_{CU}=-\frac{1}{K-1}\sum_{j=2}^{K}\sum_{r=1}^{C}\hat{\boldsymbol{y}}_{r}^{j}\log{f_{r}^{b}(\boldsymbol{x}^{j})}.
\end{equation}
By minimizing the classification loss of both the labeled and unlabeled data, i.e., $\mathcal{L}_{CL}$ and $\mathcal{L}_{CU}$, the feature extractor $b\circ{g}$ is trained to reduce its bias towards the labeled source domain. Despite that the feature extractor may still be affected by the domain-specific factors of all the source domains, we argue that the process of feature extractor debiasing could allow us to obtain more effective domain-agnostic semantic features from data, facilitating the classifier rectification process with semantic feature projection. 
Different from \cite{Liang2020DoWR}, we do not choose to optimize the classifier using the pseudo labels since it could yield adverse effects in our experiments.

\subsubsection{Classifier Rectification}
As the feature extractor is debiased, we employ the generated semantic features of the unlabeled source to filter out the domain-specific bias in the classifier. Specifically, we first project the semantic features of the unlabeled sources, i.e., ${g}(\boldsymbol{x}^{j})$, $j=2,...,K$, to the semantic features of the labeled source, i.e., $b\circ{g}(\boldsymbol{x}^{1})$, with the semantic feature projection networks $\{v_{j}\}_{j=2}^{K}$. 
To improve the class-level domain invariance, we perform conditional projection, i.e., projecting the semantic features of the unlabeled sources to the semantic features of the labeled source which are in the same class by aligning the true labels $\boldsymbol{y}^{1}$ and the pseudo labels $\{\hat{\boldsymbol{y}}^{j}\}_{j=2}^{K}$. Thus, we minimize a feature projection loss to optimize the projection networks $\{v_{j}\}_{j=2}^K$:
\begin{equation}\label{equation-fp}
    \begin{aligned}
        \mathcal{L}_{FP}=\frac{1}{K-1}\sum_{j=2}^{K}s^{j}\left(b\circ{g}\left(\boldsymbol{x}^{1}\right)
        -v_{j}\circ{g}(\boldsymbol{x}^{j}\right)^{2},
    \end{aligned}
\end{equation}
where $s^{j}={1}(\boldsymbol{y}^{1}=\hat{\boldsymbol{y}}^{j})$, i.e., if $\boldsymbol{y}^{1}=\hat{\boldsymbol{y}}^{j}$, then $s^{j}=1$, else $s^{j}=0$. 
Through semantic feature projection, the semantic invariance in data is contained in the projection semantic features $v_{j}\circ{g}(\boldsymbol{x}^{j})$. 
Then we use it to rectify/optimize the classifier $c$ by minimizing the bias filtering loss, which is the average classification loss of the projections $v_{j}\circ{g}(\boldsymbol{x}^{j})$:
\begin{equation}\label{equation-bf}
    \mathcal{L}_{BF}=-\frac{1}{K-1}\sum_{j=2}^{K}\sum_{r=1}^{C}s^{j}\boldsymbol{y}_{r}^{1}\log{f_{r}^{v_{j}}}(\boldsymbol{x}^{j}),
\end{equation}
where $f^{v_{j}}:=c\circ{w}\circ{v_{j}}\circ{g}$ outputs the softmax classification of the projections, and $f^{v_{j}}_{r}$ is the $r$-th dimension output of $f^{v_{j}}$. 
An inter-domain attention module $w$ is designed to enhance the similarities of semantic information among domains, which will be introduced in the following.
By minimizing Eq. (\ref{equation-bf}), the classifier $c$ uses invariant semantic information contained in the projections to filter out the domain-specific bias and capture invariant correlation between the features and the labels. In Section \ref{sectio-theoretical-insights}, we provide theoretical insights to make it clearer and more specific.

\begin{figure}[t]
    \centering
    \includegraphics[trim={0cm 0cm 0cm 0cm},clip,width=1\columnwidth]{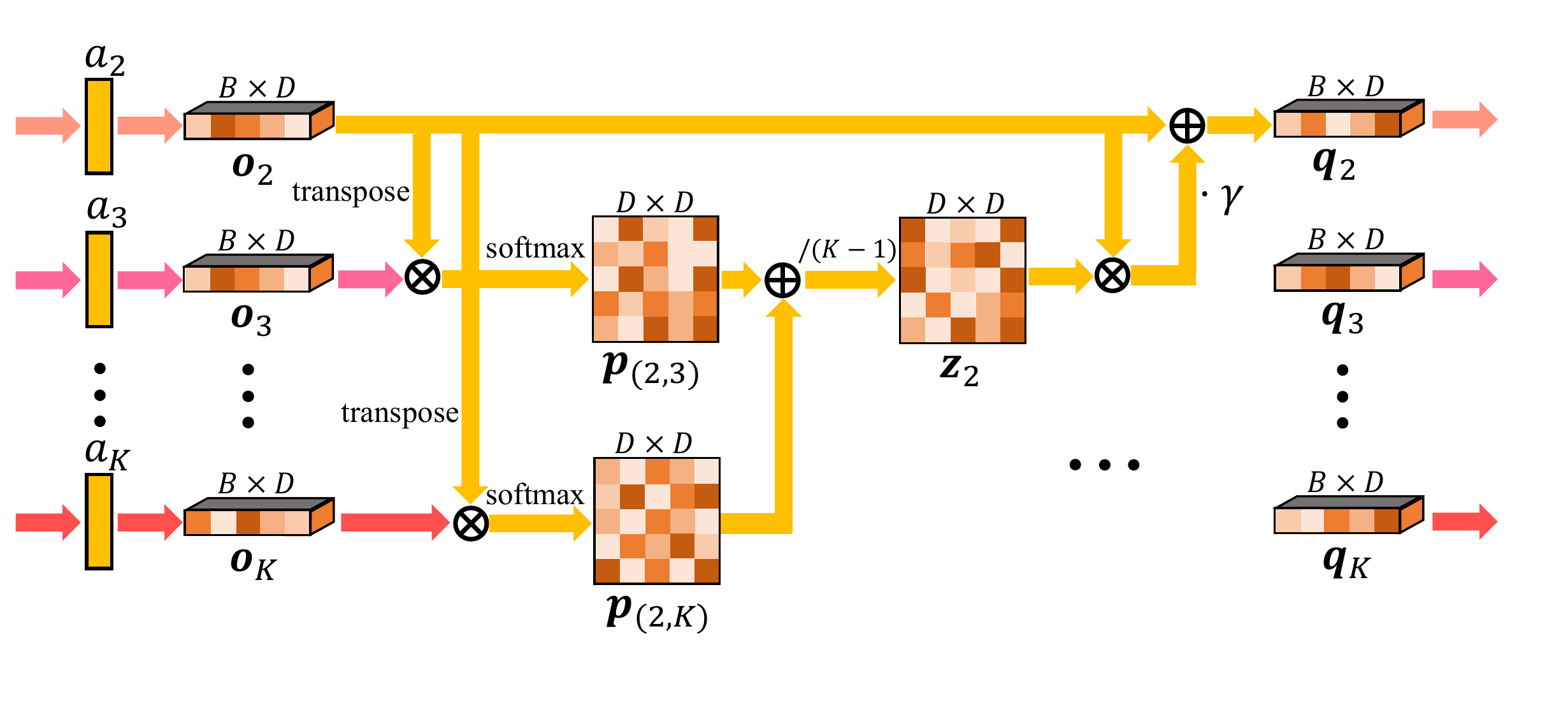} 
    \caption{The proposed inter-domain attention module. It first generates semantic feature embeddings $\{\boldsymbol{o_{j}}\}_{j=2}^{K}$ with embedding networks $\{a_{j}\}_{j=2}^{K}$, then weights them according to the inter-domain similarities and obtain weighted semantic features $\{\boldsymbol{q_{j}}\}_{j=2}^{K}$ for classifier calibrating. Domain invariance is enhanced automatically in this process.}
    \label{fig-attention}
\end{figure}

In order to further facilitate the bias filtering process, we put forward an inter-domain attention module to enhance the domain similarities as shown in Fig. \ref{fig-attention}. 
Let $B$ be the batchsize and $D$ be the semantic feature dimension, we feed the outputs of the projection networks $\{v_{j}\}_{j=2}^{K}$ with size $B\times{D}$ to embedding networks $\{a_{j}\}_{j=2}^{K}$ and get semantic feature embeddings $\{\boldsymbol{o}_{j}\}_{j=2}^{K}$ with size $B\times{D}$. Our goal is to obtain the re-weighted $\{\boldsymbol{o}_{j}\}_{j=2}^{K}$, i.e., $\{\boldsymbol{q}_{j}\}_{j=2}^{K}$, based on the aggregated inter-domain similarities among $\{\boldsymbol{o}_{j}\}_{j=2}^{K}$, i.e., $\{\boldsymbol{z}_{j}\}_{j=2}^{K}$, for more effective bias filtering and generalization boost. 

We begin by taking a domain $m$ as an example, where $m\in\{2,...,K\}$. Note that we denote $m$ as a chosen domain, and denote $j$ as the other domains used to calculate inter-domain similarities and attention. 
We first get the normalized inter-domain similarity matrices $\{\boldsymbol{p}_{(m,j)}\}_{j=2}^{K}$ of $\boldsymbol{o}_{m}$ by multiplying the transpose of $\boldsymbol{o}_{m}$ with $\{\boldsymbol{o}_{j}\}_{j=2}^{K}$:
\begin{equation}\label{equation-p}
    \boldsymbol{p}_{(m,j)}=\frac{\exp\left(\left(\boldsymbol{o}_{m}\right)^{\top}\boldsymbol{o}_{j}\right)}{\sum_{j=2}^{K}\exp\left(\left(\boldsymbol{o}_{m}\right)^{\top}\boldsymbol{o}_{j}\right)}, j=2,...,K,
\end{equation}
where $\boldsymbol{p}_{(m,j)}$ is the inter-domain similarity matrix of domain $m$ and $j$ with size $D\times{D}$ . Each position of $\boldsymbol{p}_{(m,j)}$ represents the similarities between the corresponding position of $\boldsymbol{o}_{m}$ and $\boldsymbol{o}_{j}$. Since $\{\boldsymbol{o}_{j}\}_{j=2}^{K}$ is learned from the projection of each unlabeled source to the labeled source, and $\{\boldsymbol{p}_{(m,j)}\}_{j=2}^{K}$ extract the common semantic information between $\boldsymbol{o}_{m}$ and $\{\boldsymbol{o}_{j}\}_{j=2}^{K}$, averaging $\{\boldsymbol{p}_{(m,j)}\}_{j=2}^{K}$ encourages the aggregation of the common semantic information, that is,
\begin{equation}
    \boldsymbol{z}_{m}=\frac{1}{K-1}\sum_{j=2}^{K}\boldsymbol{p}_{(m,j)}.
\end{equation}
Each position of $\boldsymbol{z}_{m}$ represents the overall response of the projection of other domains to the projection of domain $m$, which also indicates the common semantic information of each position of them. Then we get the re-weighted semantic features $\boldsymbol{q}_{m}$ by multiplying $\boldsymbol{o}_{m}$ with $\boldsymbol{z}_{m}$, and perform an element-wise sum operation with $\boldsymbol{o}_{m}$, that is,
\begin{equation}\label{equation-q}
    \boldsymbol{q}_{m}=\alpha\cdot\boldsymbol{o}_{m}\boldsymbol{z}_{m}+\boldsymbol{o}_{m},
\end{equation}
where $\alpha$ is a parameter initialized as 0 and trained to provide suitable weight. It is updated with the model parameters together (we add a small perturbation from a uniform distribution $U(0,1)$ to it to make it trained stably).
In this way, through the calculation of all the embedding semantic features $\{\boldsymbol{o}_{j}\}_{j=2}^{K}$, we can obtain all the re-weighted semantic features $\{\boldsymbol{q}_{j}\}_{j=2}^{K}$, which is re-weighted by semantic similarities among the source domains. This inter-domain attention module encourages the learning of the common semantic information of the semantic feature projection. It effectively assists the bias filtering and improve the generalization performance as verified in the experiments.

\textbf{Remark.} Note that Fig. \ref{fig-attention} is simplified. In our experiments, each network in $\{a_{j}\}_{j=2}^{K}$ is composed of three sub-networks that output the embeddings of query $\{\boldsymbol{o}_{j}^{query}\}_{j=2}^{K}$, key $\{\boldsymbol{o}_{j}^{key}\}_{j=2}^{K}$, and value $\{\boldsymbol{o}_{j}^{value}\}_{j=2}^{K}$, respectively. 
Eq. (\ref{equation-p}) is calculated with the key part of $\boldsymbol{o}_{m}$ and query part of $\boldsymbol{o}_{j}$, i.e., $\boldsymbol{o}_{m}^{key}$ and $\boldsymbol{o}_{j}^{query}$. While Eq. (\ref{equation-q}) is calculated with the value part of $\boldsymbol{o}_{m}$, i.e., $\boldsymbol{o}_{m}^{value}$.

\subsubsection{Optimization Details}\label{sec-opt}
To illustrate the optimization process clearly, we merge the optimization losses to a loss for stage 1, i.e. $\mathcal{L}_{S1}$, and a loss for stage 2, i.e. $\mathcal{L}_{S2}$, that is,
\begin{equation}
    \begin{aligned}
        \mathcal{L}_{S1}&=\mathcal{L}_{CL}\\
        \mathcal{L}_{S2}&=\lambda(\mathcal{L}_{IM}+\mathcal{L}_{CU})+\gamma(\mathcal{L}_{FP}+\mathcal{L}_{BF})
    \end{aligned}
\end{equation}
In the first stage, we initialize the model with the classification loss $\mathcal{L}_{CL}$ on the labeled source data. In the second stage, we debias the feature extractor through the classification loss $\mathcal{L}_{CU}$ on the unlabeled source data with the pseudo labels obtained via k-means clustering and the information maximization loss $\mathcal{L}_{IM}$. We rectify the classifier with the feature projection loss $\mathcal{L}_{FP}$ and the bias filtering loss $\mathcal{L}_{BF}$. $\lambda$ and $\gamma$ are the hyper-parameters for balancing the feature extractor debiasing and the classifier rectification processes. 

\textbf{The CDG task.} In the CDG task, since the groud-truth labels of all the source data are given, we directly employ them for training instead of obtaining pseudo labels through the clustering.

\subsection{Theoretical Insights}\label{sectio-theoretical-insights}
In the SLDG task, since only one source dataset is labeled, we put forward to rectify the classifier by performing the semantic feature projection. For simplicity, we denote the semantic features extracted from data $X^{j}$ as $H^{j}\in\mathbb{R}^{d_{h}}$, and let it be composed of domain-invariant factor $U\in\mathbb{R}^{d_{h}}$ and domain-specific factor/bias $L^{j}\in\mathbb{R}^{d_{h}}$, that is,
\begin{equation}\label{equation-h}
    H^{j}=(\boldsymbol{\phi}^{j})^{\top}U+(\boldsymbol{\eta}^{j})^{\top}L^{j}, j=1,...,K,
\end{equation}
where $\boldsymbol{\phi}^{j}\in\mathbb{R}^{d_{h}\times{d_{h}}}$ and $\boldsymbol{\eta}^{j}\in\mathbb{R}^{d_{h}\times{d_{h}}}$ are coefficient matrices, which may change across the domains. 

Inspired by the ability of the human in robust visual object recognition that no matter how the domain/environment changes, human can always accurately identify the class of the recognized image \citep{zhang2020acausal}. We assume that there is an invariant correlation $\beta$ between the semantic features $H^{j}$ and the corresponding labels $Y^{j}\in\mathbb{R}$, meanwhile, the labels $Y^{j}$ may also be biased by the domain-specific factor $L^{j}$, that is,
\begin{equation}\label{equation-y}
    Y^{j}=\beta^{\top}H^{j}+(\psi^{j})^{\top}L^{j}, j=1,...,K,
\end{equation}
where $\beta\in\mathbb{R}^{d_{h}}$ and $\psi^{j}\in\mathbb{R}^{d_{h}}$ are coefficient vectors. $\beta$ stays unchanged but $\psi^{j}$ changes across the domains. Note that we assume $\mathbb{E}[L^{j}]=0$ for $j=1,...,K$. The main assumption is summarized:
\begin{Assumption}\label{invariant-asp}
The semantic features $H^{j}$ and the labels $Y^{j}$ in each domain $j$ satisfy Eq. (\ref{equation-h}) and (\ref{equation-y}) respectively, where only the domain-invariant factor $U$ and the correlation $\beta$ stay unchanged across domains. The domain-specific and domain-invariant factors are pairwise independent, i.e., ${U}\perp{L^{k}}$ and $L^{j}\perp{L^{k}}$ for $j,k\in\{1,...,K\}$ and $j\neq{k}$.
\end{Assumption}
Our goal is to identify the latent correlation $\beta$ between the features and the labels. Let $m$ and $n$ be an unlabeled and a labeled source domain, respectively. We first project the semantic features $H^{m}$ of the unlabeled source data to the semantic features $H^{n}$ of the labeled source data with a mapping matrix $\boldsymbol{\mu}\in\mathbb{R}^{d_{h}\times{d_{h}}}$, that is, $\hat{\boldsymbol{\mu}}=\mathbb{E}[H^{m}(H^{m})^{\top}]^{-1}\mathbb{E}[H^{m}(H^{n})^{\top}]$. Then we use the projection semantic features, i.e., $\hat{H}^{n}=\hat{\boldsymbol{\mu}}^{\top}H^{m}$, to fit the labels $Y^{n}$ of the labeled source and estimate the correlation $\hat{\beta}=\mathbb{E}[\hat{H}^{n}(\hat{H}^{n})^{\top}]^{-1}[\hat{H}^{n}(Y^{n})^{\top}]$, i.e., classifier rectification. 
We have the theorem: 
\begin{Theorem}\label{beta}
    Suppose there are $n$ samples from each domain, then $\hat{\beta}$ is a consistent and unbiased estimator of the true correlation $\beta$, i.e., $\hat{\beta}=\beta+O_{p}\left(\frac{1}{\sqrt{n}}\right)$ and $\mathbb{E}[\hat{\beta}]=\beta$.
\end{Theorem}
\emph{Proof.}
Assume that we sample $n$ examples from each domain. Let $\*H^{m}\in\mathbb{R}^{n\times{d_{h}}}$ be the matrix where $i$th row is the observation $\boldsymbol{h}_{i}^{m}\in\mathbb{R}^{d_{h}}$ of $H^{m}$, and other symbols are similarly defined.
The first step is to regress $\mathbf{H}^{n}$ on $\mathbf{H}^{m}$, i.e., $\hat\mu = \big( (\mathbf{H}^{m})^{\top} \mathbf{H}^{m} \big)^{-1} (\mathbf{H}^{m})^{\top} \mathbf{H}^{n}$.
The second step is to regress $\mathbf{Y}^{n}$ on $\hat{\mathbf{H}}^n = \mathbf{H}^{m} \hat \mu$, i.e., $\hat{\beta} = \big( (\hat{\mathbf{H}}^n)^{\top} \hat{\mathbf{H}}^n \big)^{-1} (\hat{\mathbf{H}}^n)^{\top} \mathbf{Y}^{n}$.

By Assumption \ref{invariant-asp}, we have
\begin{equation}\label{equation-xt-fs}
    \begin{aligned}
        \frac{1}{n} (\mathbf{H}^{m})^{\top} \mathbf{L}^{n}
        =& \frac{1}{n} \big( \*U \boldsymbol{\phi}^{m} + \*L^{m} \boldsymbol{\eta}^{m}\big)^\top\mathbf{L}^{n}\\
        =& O_p\bigg( \frac{1}{\sqrt{n}} \bigg).
    \end{aligned}
\end{equation}
\begin{equation}\label{equation-xs-xt}
    \begin{aligned}
        \frac{1}{n} (\mathbf{H}^{n})^{\top} \mathbf{H}^{m}
        =& \frac{1}{n}  \big( \*U \boldsymbol{\phi}^{n} + \*L^{n} \boldsymbol{\eta}^{n}\big)^\top\cdot\big( \*U \boldsymbol{\phi}^{m} + \*L^{m} \boldsymbol{\eta}^{m}\big) \\
        =& \frac{1}{n} (\boldsymbol{\phi}^{n})^\top\*U^\top\*U \boldsymbol{\phi}^{m} + O_p\bigg( \frac{1}{\sqrt{n}} \bigg),
    \end{aligned}
\end{equation}
\begin{equation}\label{equation-xt-xt}
    \begin{aligned}
    &\frac{1}{n} (\mathbf{H}^{m})^{\top} \mathbf{H}^{m}\\ 
    =& \frac{1}{n}  \big( \*U \boldsymbol{\phi}^{m} + \*L^{m} \boldsymbol{\eta}^{m}\big)^\top\cdot \big( \*U \boldsymbol{\phi}^{m} + \*L^{m} \boldsymbol{\eta}^{m}\big) \\
    =& \frac{1}{n} \left((\boldsymbol{\phi}^{m})^\top\*U^\top\*U \boldsymbol{\phi}^{m}+(\boldsymbol{\eta}^{m})^{\top}(\*L^{m})^\top\*L^{m}\boldsymbol{\eta}^{m}\right.\\
    &\left.+O_p\bigg( \frac{1}{\sqrt{n}} \bigg)\right).
\end{aligned}
\end{equation}

Suppose the minimum eigenvalue of $(\boldsymbol{\phi}^{m})^\top\cdot \+E[UU^\top] \cdot \boldsymbol{\phi}^{m}$ is bounded away from 0, we have 
\begin{equation}\label{equation-reverse}
    \begin{aligned}
    & \bigg(\frac{1}{n}(\boldsymbol{\phi}^{m})^\top\*U^\top\*U \boldsymbol{\phi}^{m} + O_p\bigg( \frac{1}{\sqrt{n}} \bigg) \bigg)^\I \\
    =& \Big((\boldsymbol{\phi}^{m})^\top\cdot \+E[UU^\top] \cdot \boldsymbol{\phi}^{m} \Big)^\I + O_p\bigg( \frac{1}{\sqrt{n}}  \bigg).
    \end{aligned}
\end{equation}
Since $(\boldsymbol{\eta}^{m})^{\top}(\*L^{m})^{\top}\*L^{m}\boldsymbol{\eta}^{m}/n$ is positive semidefinite matrices. Hence, the minimum eigenvalue of $(\boldsymbol{\phi}^{m})^\top\cdot \+E[U(U)^\top] \cdot \boldsymbol{\phi}^{m} + (\boldsymbol{\eta}^{m})^\top\cdot \+E[L^{m}(L^{m})^\top] \cdot \boldsymbol{\eta}^{m}$ is bounded away from 0, then
\begin{equation}\label{equation-reverse-fivt-ft-ext}
    \begin{aligned}
        &\bigg(\frac{1}{n} \bigg((\boldsymbol{\phi}^{m})^\top\*U^\top\*U \boldsymbol{\phi}^{m}+(\boldsymbol{\eta}^{m})^{\top}(\*L^{m})^{\top}\*L^{m}\boldsymbol{\eta}^{m}\\
        &+ O_p\bigg( \frac{1}{\sqrt{n}} \bigg)\bigg)\bigg)^{-1}\\
        =&\big((\boldsymbol{\phi}^{m})^\top\cdot \+E[UU^\top] \cdot \boldsymbol{\phi}^{m}\\
        &+(\boldsymbol{\eta}^{m})^\top\cdot \+E[L^{m}(L^{m})^\top] \cdot \boldsymbol{\eta}^{m}\big)^{-1}+O_p\bigg( \frac{1}{\sqrt{n}} \bigg).
    \end{aligned}
\end{equation}
Therefore, by Eq. (\ref{equation-xt-fs}-\ref{equation-reverse-fivt-ft-ext}), we have
\begin{align*}
    \hat{\beta} =& \left(\left(\hat{\mathbf{H}}^n\right)^{\top} \hat{\mathbf{H}}^n \right)^{-1} (\hat{\mathbf{H}}^n)^{\top} \mathbf{Y}^{n} \\
    =& \left( (\mathbf{H}^{n})^{\top} \mathbf{H}^{m} \big( (\mathbf{H}^{m})^{\top} \mathbf{H}^{m} \big)^{-1}\left(\mathbf{H}^{m}\right)^{\top} \mathbf{H}^{n}\right)^\I  \\ 
    & \cdot (\mathbf{H}^{n})^{\top} \mathbf{H}^{m} \big( (\mathbf{H}^{m})^{\top} \mathbf{H}^{m} \big)^{-1} (\mathbf{H}^{m})^{\top}\\
    &\cdot \big(\mathbf{H}^{n} \beta+\mathbf{L}^{n}\psi^{n} \big)  \\
    =&\beta + O_p\bigg( \frac{1}{\sqrt{n}} \bigg).
\end{align*}
We then have $\mathbb{E}[\hat{\beta}]=\beta$.

\begin{figure*}[t]
    \centering
    \includegraphics[trim={0cm 0cm 0cm 0cm},clip,width=0.58\columnwidth]{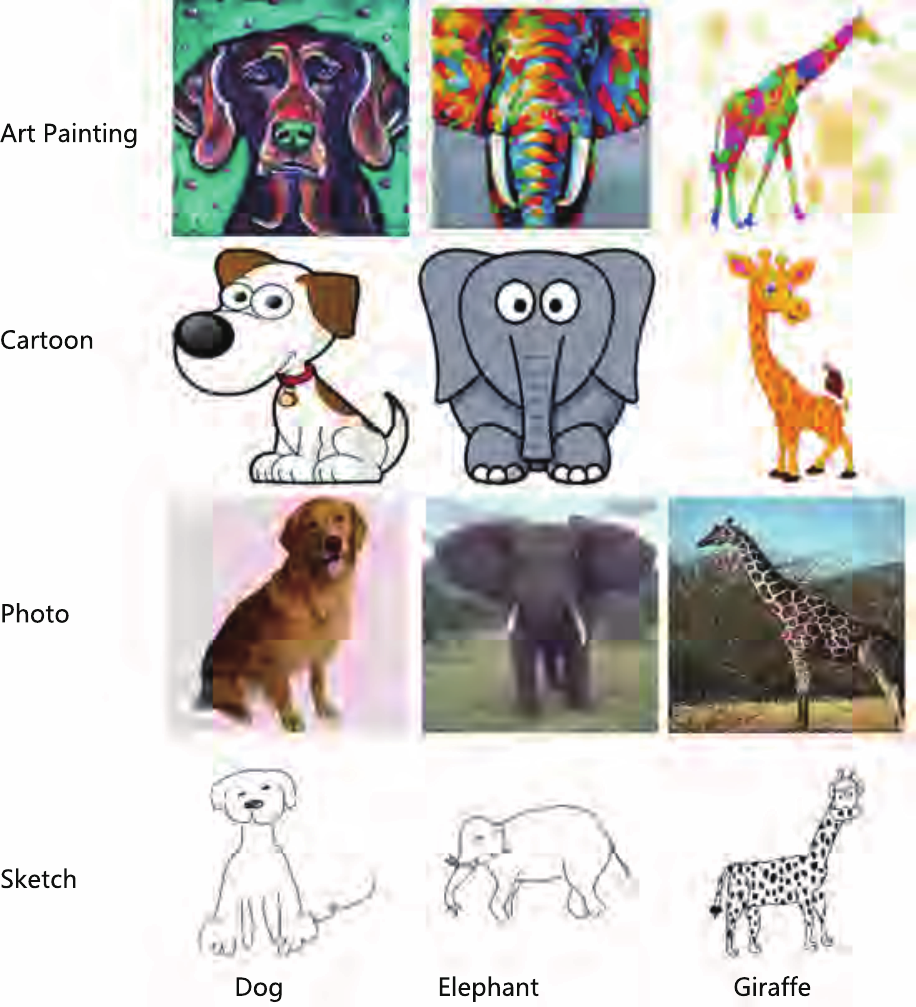}
    \includegraphics[trim={0cm 0cm 0cm 0cm},clip,width=0.61\columnwidth]{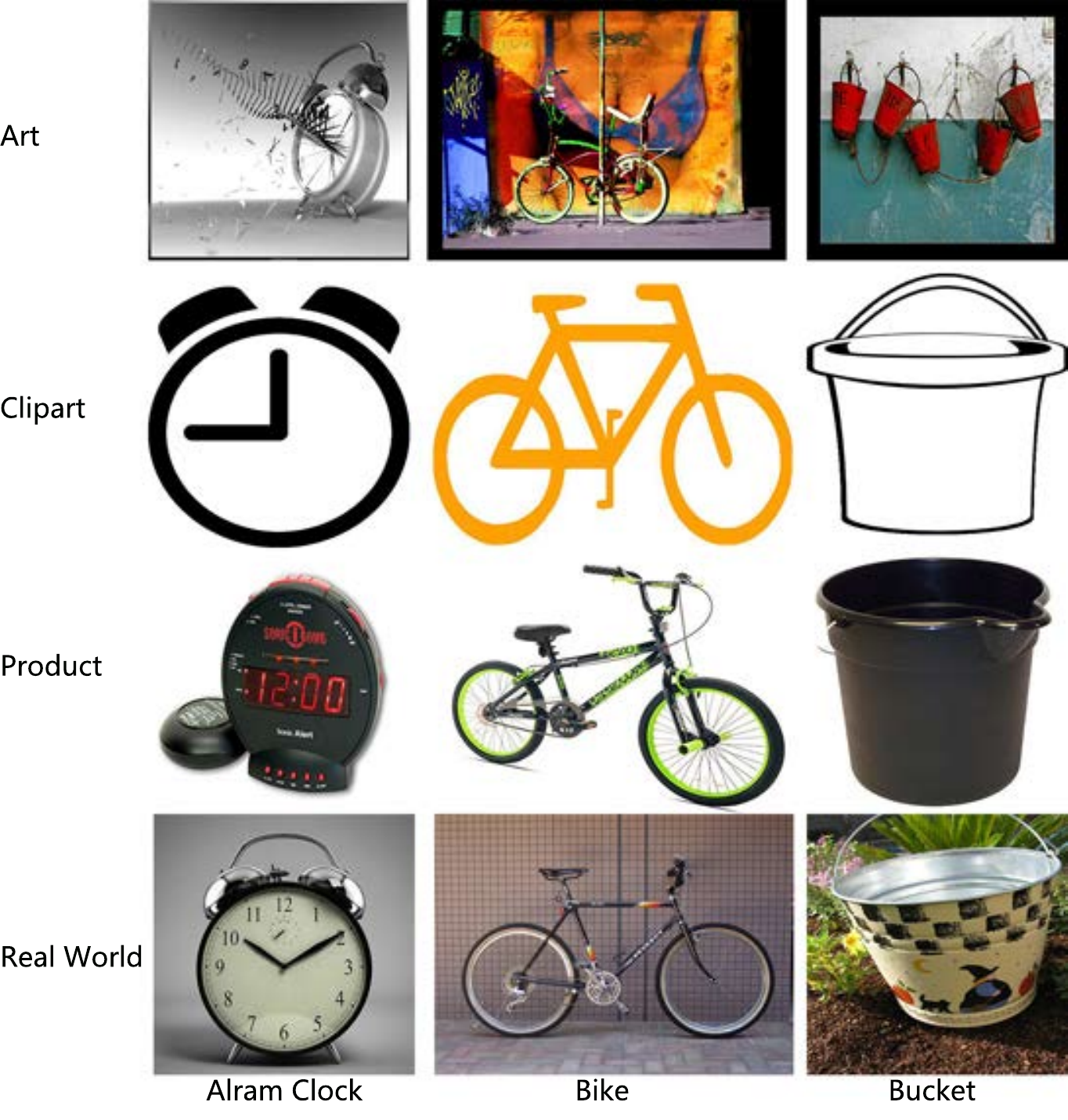}
    \includegraphics[trim={0cm 0cm 0cm 0cm},clip,width=0.76\columnwidth]{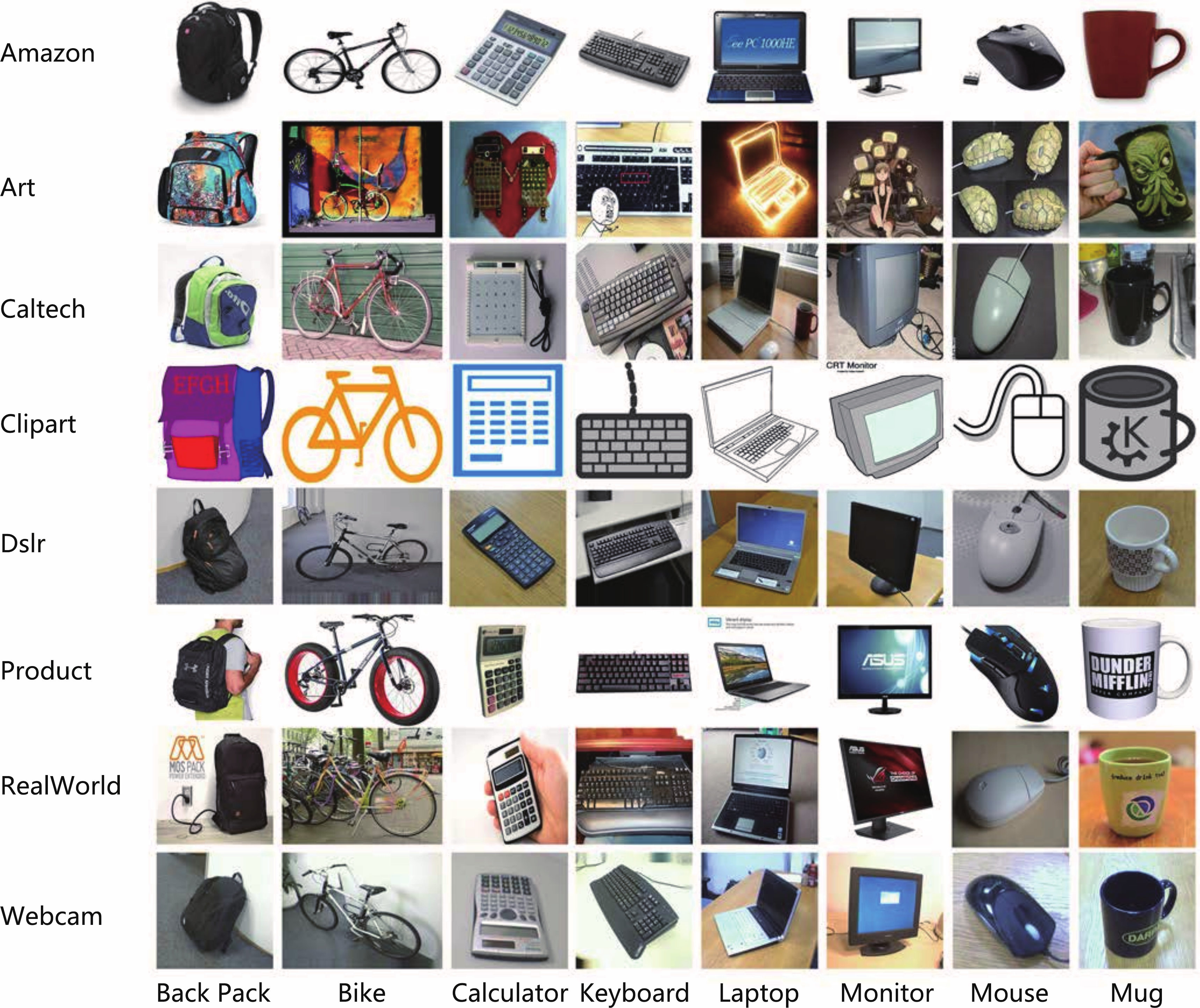}
    \caption{Example images of the datasets. \textbf{Left}: PACS dataset \citep{li2017deeper} with four domains, i.e., Art Painting (Ar), Cartoon (Ca), Photo (Ph), and Sketch (Sk). \textbf{Middle}: Office-Home dataset \citep{venkateswara2017deep} with four domains, i.e., Art (Ar), Clipart (Cl), Product (Pr), and Real World (Rw). \textbf{Right}: our Office-Caltech-Home dataset with eight domains, i.e., Amazon (Am), Art (Ar), Caltech (Ca), Clipart (Cl), Dslr (Ds), Product (Pr), Real World (Rw), and Webcam (We).}
    \label{fig-datasets}
\end{figure*}

Theorem \ref{beta} indicates that we can use the semantic features of the unlabeled source to filter out the domain-specific factor/bias of the labeled source and capture the domain-invariant correlation $\beta$ for more stable domain generalization. Since our theoretical analyses are based on the linear setting, we design the inter-domain attention module to further improve the bias filtering process by learning from the similarities among domains. In this way, the common parts of the features of the unlabeled source domains are enhanced, which helps to further remove the bias and boost the generalization performance.

\section{Experiments}
We first implement experiments for the introduced Single Labeled Domain Generalization (SLDG) task. We compare our method DSBF with the standard Supervised Learning (SL) algorithm as well as the state-of-the-art algorithms of Semi-Supervised Learning (SSL), Unsupervised Domain Adaptation (UDA), Multi-Target Domain Adaptation (MTDA), and Domain Generalization (DG). 
To further testify the performance of our method DSBF in domain-specific bias filtering, we then include the comparison with the other DG methods for the Conventional Domain Generalization (CDG) task, where the labels of all the source datasets are provided.

\subsection{Setup}
\textbf{Benchmark datasets.}
We first adopt two popular benchmark datasets. 
One is \textbf{PACS} \citep{li2017deeper} that contains 9,991 images from 7 classes in 4 domains, i.e., Artpaint (\textit{Ar}), Cartoon (\textit{Ca}), Sketch (\textit{Sk}), and Photo (\textit{Ph}). Another one is \textbf{Office-Home} \citep{venkateswara2017deep} that consists about 15,500 images of 65 categories over 4 domains, i.e., Art (\textit{Ar}), Clipart (\textit{Cl}), Product (\textit{Pr}), and Real-World (\textit{Rw}). 
We then perform experiments on a more challenging large-scale dataset called \textbf{DomainNet} \citep{peng2019moment}. By following \citep{zhou2021domainadaptive}, we use four representative domains, i.e., Clipart (\textit{Cl}), Painting (\textit{Pa}), Real (\textit{Re}), and Sketch (\textit{Sk}), for the experiments. 
In order to further evaluate the performance under the scenarios of more unlabeled source datasets, we construct a new dataset \textbf{Office-Caltech-Home}. Specifically, we choose the common classes from Office-Caltech \citep{gong2012geodesic} and Office-Home \citep{venkateswara2017deep} datasets, i.e., backpack, bike, calculator, keyboard, laptop (computer), monitor, mouse, mug, and merge the two datasets to be a new dataset Office-Caltech-Home that has 4,266 images of 8 classes in 8 domains, i.e., Amazon (\textit{Am}), Webcam (\textit{We}), DSLR (\textit{Ds}), Caltech (\textit{Ca}), Art (\textit{Ar}), Clipart (\textit{Cl}), Product (\textit{Pr}), and Real-World (\textit{Rw}). We discard DSLR since it only has few images. We use the rest 7 domains with 4,145 images. Example images are shown in Fig. \ref{fig-datasets}.

\textbf{Baseline methods.}
For the experiments of the SLDG task, multiple source datasets are used for model training but only one of them is labeled. 
The first baseline method is the standard Supervised Learning (SL). ERM \citep{vapnik1992principles} is employed that minimizes the empirical risk, i.e., cross-entropy loss of classification, on the labeled source dataset. 
For Semi-Supervised Learning (SSL), both the labeled source dataset and the mixture of the unlabeled source datasets are utilized. We conduct two state-of-the-art SSL methods, i.e., Mean Teacher \citep{Tarvainen2017MeanTA} and FixMatch \citep{SohnBCZZRCKL20}, their strategies are related to knowledge distillation \citep{hinton2015distilling} and data augmentation, respectively.
We also compare DSBF with Unsupervised Domain Adaptation (UDA), where the labeled source dataset and the mixture of the unlabeled source datasets are used as the source dataset and the unlabeled target dataset in the UDA task, respectively. Several representative UDA methods are considered, i.e., DAN \citep{long2015learning}, MCD \citep{saito2018maximum}, and MDD \citep{Zhang2019BridgingTA}.
Besides, Multi-Target Domain Adaptation (MTDA) is considered to use the labeled source dataset and multiple unlabeled source datasets as the labeled source dataset and multiple unlabeled target datasets, respectively. We employ the state-of-the-art MTDA methods, i.e., BTDA \citep{Chen2019BlendingTargetDA} and OCDA \citep{Liu2020OpenCD} as the baselines. 
Since only one labeled dataset can be utilized in the SLDG task, we compare DSBF with the DG methods which can be extended to this task, including data augmentation based methods JiGen \citep{Carlucci2019DomainGB} and GUD \citep{Volpi2018GeneralizingTU}, a training heuristics method RSC \citep{HuangWXH20}, and a single domain method M-ADA \citep{qiao2020learning}. 
These works have been introduced in Section \ref{sec:rel}.

\begin{sidewaystable*}
\caption{Classification accuracy (\%) for the Single Labeled Domain Generalization (SLDG) task on PACS dataset. A$\rightarrow$B represents using A, B, and other domains as the labeled source, target, and unlabeled source domains, respectively. The best results are emphasized in bold.}
    \label{table-dgsls-pacs}
\scalebox{0.6}[0.6]{
\begin{tabular}{l|c|cccccccccccc|c}
\toprule
Methods & Type & Ca$\rightarrow$Ar & Sk$\rightarrow$Ar & Ph$\rightarrow$Ar & Ar$\rightarrow$Ca & Sk$\rightarrow$Ca & Ph$\rightarrow$Ca & Ar$\rightarrow$Sk & Ca$\rightarrow$Sk & Ph$\rightarrow$Sk & Ar$\rightarrow$Ph & Ca$\rightarrow$Ph & Sk$\rightarrow$Ph & Avg.\\
\midrule
ERM \citep{vapnik1992principles} & SL & 65.25$\pm$2.20 & 25.88$\pm$2.53 & \textbf{66.65}$\pm$2.80 & 57.89$\pm$1.09 & 53.37$\pm$4.59 & 25.53$\pm$1.27 & 51.34$\pm$1.85 & 65.00$\pm$3.07 & 30.52$\pm$0.93 & \textbf{97.01}$\pm$3.67 & 87.13$\pm$0.29 & 37.72$\pm$1.98 & 55.27$\pm$0.74 \\
\midrule
Mean Teacher \citep{Tarvainen2017MeanTA} & \multirow{2}{*}{SSL} & 30.66$\pm$1.83 & 17.43$\pm$2.75 & 28.61$\pm$0.80 & 33.36$\pm$1.44 & 21.50$\pm$0.91 & 28.16$\pm$1.53 & 25.78$\pm$2.50 & 38.20$\pm$2.51 & 29.07$\pm$1.77 & 42.04$\pm$2.00 & 37.07$\pm$1.30 & 16.71$\pm$2.45 & 29.05$\pm$0.68 \\
FixMatch \citep{SohnBCZZRCKL20} & & 30.03$\pm$5.41 & 14.45$\pm$1.69 & 22.85$\pm$0.31 & 31.66$\pm$2.13 & 24.79$\pm$1.50 & 25.64$\pm$3.74 & 19.64$\pm$1.46 & 39.07$\pm$1.28 & 29.68$\pm$1.04 & 25.87$\pm$0.98 & 33.95$\pm$0.83 & 14.31$\pm$1.81 & 26.00$\pm$1.03 \\
\midrule
DAN \citep{long2015learning} & \multirow{4}{*}{UDA} & 62.8$\pm$3.0 & 46.9$\pm$3.3 & 58.3$\pm$0.8 & 68.3$\pm$0.8 & 62.5$\pm$3.9 & 48.2$\pm$1.0 & 55.1$\pm$2.7 & 54.8$\pm$0.6 & 31.3$\pm$2.4 & 93.3$\pm$1.9 & 85.0$\pm$0.7 & 58.9$\pm$2.9 & 60.5$\pm$0.8 \\
MCD \citep{saito2018maximum} & & 72.8$\pm$0.5 & 29.7$\pm$1.3 & 66.0$\pm$3.6 & \textbf{69.7}$\pm$1.0 & 61.1$\pm$4.2 & 43.6$\pm$4.2 & 49.1$\pm$2.6 & 61.6$\pm$3.6 & 25.2$\pm$2.0 & 95.9$\pm$3.2 & 83.5$\pm$4.6 & 44.2$\pm$1.0 & 58.5$\pm$0.4 \\
MDD \citep{Zhang2019BridgingTA} & & 71.7$\pm$1.4 & 40.2$\pm$3.0 & 61.9$\pm$3.0 & 67.0$\pm$1.1 & 58.1$\pm$2.2 & \textbf{62.8}$\pm$6.3 & 44.8$\pm$2.3 & 57.1$\pm$3.8 & 33.5$\pm$1.3 & 95.9$\pm$2.1 & 85.1$\pm$1.1 & 55.0$\pm$3.7 & 61.1$\pm$0.9 \\
SHOT \citep{Liang2020DoWR} & & 71.63$\pm$1.07 & 58.37$\pm$1.89 & 49.10$\pm$1.27 & 66.30$\pm$1.16 & 44.50$\pm$1.87 & 57.30$\pm$3.26 & 63.70$\pm$0.31 & 61.40$\pm$1.23 & 58.50$\pm$1.88 & 82.20$\pm$2.81 & 85.20$\pm$0.26 & 62.10$\pm$2.74 & 63.36$\pm$0.29 \\
\midrule
OCDA \citep{Liu2020OpenCD} & \multirow{2}{*}{MTDA} & 17.33$\pm$4.15 & 18.51$\pm$4.34 & 17.45$\pm$0.87 & 26.07$\pm$4.16 & 15.78$\pm$0.32 & 19.71$\pm$5.45 & 16.47$\pm$1.14 & 19.67$\pm$0.46 & 24.31$\pm$2.50 & 26.43$\pm$2.69 & 24.07$\pm$0.66 & 11.32$\pm$3.58 & 19.76$\pm$0.43 \\
BTDA \citep{Chen2019BlendingTargetDA} & & 63.58$\pm$3.64 & 47.14$\pm$5.65 & 54.95$\pm$1.27 & 59.15$\pm$3.36 & 35.74$\pm$2.71 & 22.32$\pm$0.17 & \textbf{70.90}$\pm$2.20 & 64.79$\pm$0.97 & 31.78$\pm$1.69 & 42.76$\pm$1.66 & 60.93$\pm$0.22 & 48.76$\pm$3.10 & 50.24$\pm$0.45 \\
\midrule
GUD \citep{Volpi2018GeneralizingTU} & \multirow{4}{*}{DG} & 68.12$\pm$0.94 & 22.75$\pm$2.96 & 66.06$\pm$1.06 & 68.17$\pm$0.63 & 34.68$\pm$1.00 & 26.11$\pm$3.88 & 64.27$\pm$2.01 & 68.44$\pm$0.19 & 52.48$\pm$0.26 & 95.57$\pm$1.53 & 82.75$\pm$0.69 & 36.53$\pm$4.93 & 57.16$\pm$0.65 \\
JiGen \citep{Carlucci2019DomainGB} & & 66.60$\pm$1.36 & 23.68$\pm$0.97 & 64.36$\pm$0.48 & 53.91$\pm$1.61 & 33.45$\pm$0.67 & 26.58$\pm$0.82 & 50.24$\pm$1.87 & 62.76$\pm$1.48 & 28.58$\pm$2.30 & 95.75$\pm$0.36 & 84.31$\pm$0.31 & 30.24$\pm$1.86 & 51.71$\pm$0.11 \\
RSC \citep{HuangWXH20} & & 64.79$\pm$1.62 & 53.03$\pm$2.66 & 66.50$\pm$3.53 & 67.15$\pm$1.97 & \textbf{66.64}$\pm$3.89 & 26.58$\pm$1.10 & 55.46$\pm$1.32 & \textbf{73.96}$\pm$5.88 & 44.06$\pm$2.02 & 94.79$\pm$0.08 & 82.10$\pm$1.62 & 47.25$\pm$1.56 & 61.86$\pm$0.36 \\
M-ADA \citep{qiao2020learning} & & 55.18$\pm$2.77 & 24.41$\pm$4.71 & 55.71$\pm$2.82 & 62.33$\pm$2.87 & 58.36$\pm$5.28 & 35.92$\pm$3.96 & 51.95$\pm$4.36 & 72.28$\pm$0.99 & 31.15$\pm$3.44 & 84.43$\pm$2.20 & 71.98$\pm$2.15 & 34.92$\pm$3.09 & 53.22$\pm$0.59 \\
\midrule
\textbf{DSBF} & SLDG & \textbf{73.14}$\pm$0.71 & \textbf{64.97}$\pm$5.10 & 52.93$\pm$2.28 & 66.88$\pm$1.38 & 45.72$\pm$2.86 & 55.72$\pm$2.07 & 68.54$\pm$0.59 & 68.86$\pm$1.61 & \textbf{64.76}$\pm$2.79 & 87.26$\pm$1.23 & \textbf{89.28}$\pm$1.44 & \textbf{67.40}$\pm$3.73 & \textbf{67.12}$\pm$0.60 \\
\bottomrule
\end{tabular}}
\end{sidewaystable*}

\begin{sidewaystable*}
\caption{Classification accuracy (\%) for the Single Labeled Domain Generalization (SLDG) task on Office-Home dataset. A$\rightarrow$B represents using A, B, and other domains as the labeled source, target, and unlabeled source domains, respectively. The best results are emphasized in bold.} 
    \label{table-dgsls-home}
\scalebox{0.6}[0.6]{
\begin{tabular}{l|c|cccccccccccc|c}
\toprule
Methods & Type & Cl$\rightarrow$Ar & Pr$\rightarrow$Ar & Rw$\rightarrow$Ar & Ar$\rightarrow$Cl & Pr$\rightarrow$Cl & Rw$\rightarrow$Cl & Ar$\rightarrow$Pr & Cl$\rightarrow$Pr & Rw$\rightarrow$Pr & Ar$\rightarrow$Rw & Cl$\rightarrow$Rw & Pr$\rightarrow$Rw & Avg. \\
\midrule
ERM \citep{vapnik1992principles} & SL & 44.39$\pm$1.66 & 42.65$\pm$4.19 & \textbf{58.12}$\pm$3.17 & 38.83$\pm$0.97 & 33.96$\pm$1.20 & 40.92$\pm$0.90 & 58.14$\pm$5.34 & 56.84$\pm$3.88 & \textbf{74.25}$\pm$0.57 & 67.84$\pm$2.41 & 59.61$\pm$8.46 & 65.43$\pm$0.14 & 53.42$\pm$0.71 \\
\midrule
Mean Teacher \citep{Tarvainen2017MeanTA} & \multirow{2}{*}{SSL} & 8.41$\pm$3.68 & 14.70$\pm$5.11 & 2.84$\pm$1.75 & 8.96$\pm$3.59 & 12.41$\pm$1.22 & 13.56$\pm$0.43 & 2.23$\pm$0.35 & 10.90$\pm$1.07 & 29.29$\pm$4.87 & 4.35$\pm$0.64 & 6.34$\pm$0.80 & 18.56$\pm$0.09 & 11.05$\pm$0.84 \\
FixMatch \citep{SohnBCZZRCKL20} & & 8.94$\pm$1.46 & 7.50$\pm$0.33 & 3.13$\pm$1.06 & 8.18$\pm$0.55 & 13.13$\pm$1.07 & 16.91$\pm$1.53 & 7.16$\pm$6.55 & 14.69$\pm$2.81 & 20.01$\pm$0.24 & 2.27$\pm$0.57 & 4.36$\pm$3.87 & 21.25$\pm$2.94 & 10.63$\pm$0.80 \\
\midrule
DAN \citep{long2015learning} & \multirow{3}{*}{UDA} & 46.6$\pm$2.2 & 43.8$\pm$6.1 & 58.1$\pm$2.4 & 38.3$\pm$3.1 & 33.8$\pm$0.4 & 41.9$\pm$2.3 & 58.8$\pm$2.6 & 57.9$\pm$1.1 & 73.0$\pm$3.3 & 66.6$\pm$0.4 & 59.5$\pm$1.7 & 65.2$\pm$0.6 & 53.6$\pm$0.8 \\
MCD \citep{saito2018maximum} & & 42.3$\pm$2.3 & 42.6$\pm$0.9 & 58.1$\pm$4.1 & 35.7$\pm$2.7 & 32.3$\pm$4.7 & 39.1$\pm$2.9 & 56.3$\pm$3.3 & 53.7$\pm$4.2 & 72.6$\pm$1.8 & 65.5$\pm$0.9 & 55.4$\pm$3.0 & 64.8$\pm$3.4 & 51.5$\pm$0.8 \\
MDD \citep{Zhang2019BridgingTA} & & 45.9$\pm$0.2 & 47.4$\pm$0.3 & 56.7$\pm$1.8 & 39.4$\pm$2.7 & 34.6$\pm$2.6 & 42.9$\pm$2.4 & 59.6$\pm$1.1 & 59.1$\pm$1.9 & 72.8$\pm$0.4 & 68.1$\pm$4.6 & 61.3$\pm$1.3 & 65.5$\pm$1.1 & 54.4$\pm$0.3 \\
SHOT \citep{Liang2020DoWR} & & 51.30$\pm$4.32 & 48.73$\pm$1.53 & 48.04$\pm$0.40 & 39.98$\pm$1.86 & 37.09$\pm$2.73 & 39.89$\pm$2.79 & \textbf{63.08}$\pm$7.56 & 60.46$\pm$0.69 & 63.05$\pm$1.60 & 62.47$\pm$3.02 & 63.44$\pm$0.44 & 60.73$\pm$3.51 & 53.19$\pm$0.61 \\
\midrule
OCDA \citep{Liu2020OpenCD} & \multirow{2}{*}{MTDA} & 13.26$\pm$1.92 & 11.58$\pm$1.97 & 20.60$\pm$0.29 & 13.57$\pm$1.59 & 25.40$\pm$0.77 & 12.69$\pm$0.67 & 11.04$\pm$2.10 & 24.10$\pm$1.91 & 11.97$\pm$2.35 & 13.34$\pm$2.51 & 18.46$\pm$2.37 & 11.76$\pm$5.51 & 15.65$\pm$0.99 \\
BTDA \citep{Chen2019BlendingTargetDA} & & 39.27$\pm$1.90 & \textbf{59.97}$\pm$5.07 & 50.57$\pm$3.08 & 42.37$\pm$0.81 & \textbf{59.22}$\pm$0.71 & \textbf{57.79}$\pm$0.25 & 43.95$\pm$1.11 & 48.39$\pm$9.33 & 51.84$\pm$2.30 & 40.14$\pm$0.57 & 47.74$\pm$0.17 & 58.50$\pm$0.53 & 49.98$\pm$0.39 \\
\midrule
GUD \citep{Volpi2018GeneralizingTU} & \multirow{4}{*}{DG} & 42.31$\pm$0.46 & 31.20$\pm$1.02 & 53.03$\pm$3.00 & 38.92$\pm$3.89 & 35.46$\pm$1.33 & 43.89$\pm$1.40 & 51.79$\pm$1.14 & 50.10$\pm$2.06 & 71.71$\pm$1.94 & 61.56$\pm$1.39 & 53.22$\pm$2.00 & 60.22$\pm$0.93 & 49.45$\pm$0.51 \\
JiGen \citep{Carlucci2019DomainGB} &  & 41.62$\pm$1.92 & 38.20$\pm$3.11 & 54.31$\pm$4.71 & 36.20$\pm$0.26 & 36.08$\pm$0.94 & 42.29$\pm$3.73 & 44.78$\pm$1.31 & 53.57$\pm$5.81 & 69.63$\pm$3.91 & 55.57$\pm$1.00 & 54.97$\pm$0.24 & 62.54$\pm$1.94 & 49.15$\pm$0.33 \\
RSC \citep{HuangWXH20} & & 40.23$\pm$0.28 & 37.58$\pm$5.06 & 55.50$\pm$2.30 & 39.54$\pm$0.23 & 38.35$\pm$0.72 & 46.94$\pm$1.98 & 49.72$\pm$3.14 & 52.29$\pm$6.92 & 72.89$\pm$1.33 & 63.16$\pm$3.51 & 54.88$\pm$1.30 & 57.44$\pm$2.20 & 50.71$\pm$0.78 \\
M-ADA \citep{qiao2020learning} & & 30.08$\pm$1.24 & 23.61$\pm$2.65 & 44.21$\pm$1.55 & 37.55$\pm$1.01 & 33.68$\pm$4.60 & 44.15$\pm$0.40 & 43.10$\pm$0.80 & 30.39$\pm$1.40 & 63.05$\pm$0.86 & 53.22$\pm$1.11 & 44.50$\pm$1.83 & 49.83$\pm$5.35 & 41.45$\pm$0.44 \\
\midrule
\textbf{DSBF} & SLDG & \textbf{51.37}$\pm$0.92 & 50.30$\pm$0.99 & 56.90$\pm$0.23 & \textbf{43.81}$\pm$0.01 & 37.18$\pm$0.34 & 42.43$\pm$0.28 & 61.19$\pm$0.59 & \textbf{61.23}$\pm$1.29 & 71.71$\pm$1.12 & \textbf{68.58}$\pm$0.61 & \textbf{64.30}$\pm$0.53 & \textbf{67.84}$\pm$0.57 & \textbf{56.40}$\pm$0.12 \\
\bottomrule
\end{tabular}}
\end{sidewaystable*}

\begin{sidewaystable*}
\caption{Classification accuracy (\%) for the Single Labeled Domain Generalization (SLDG) task on DomainNet dataset. A$\rightarrow$B represents using A, B, and other domains as the labeled source, target, and unlabeled source domains, respectively. The best results are emphasized in bold.} 
    \label{table-dgsls-domainnet}
\scalebox{0.6}[0.6]{
\begin{tabular}{l|c|cccccccccccc|c}
\toprule
Methods & Type & Pa$\rightarrow$Cl & Re$\rightarrow$Cl & Sk$\rightarrow$Cl & Cl$\rightarrow$Pa & Re$\rightarrow$Pa & Sk$\rightarrow$Pa & Cl$\rightarrow$Re & Pa$\rightarrow$Re & Sk$\rightarrow$Re & Cl$\rightarrow$Sk & Pa$\rightarrow$Sk & Re$\rightarrow$Sk & Avg. \\
\midrule
ERM \citep{vapnik1992principles} & SL & 30.93$\pm$1.50 & 34.73$\pm$1.45 & 40.98$\pm$1.39 & 29.42$\pm$1.34 & 36.33$\pm$0.76 & 32.95$\pm$0.53 & 40.88$\pm$0.76 & 47.04$\pm$0.76 & 40.89$\pm$1.20 & 28.99$\pm$1.49 & 24.56$\pm$1.44 & 25.73$\pm$1.28 & 34.45$\pm$0.61 \\
\midrule
Mean Teacher \citep{Tarvainen2017MeanTA} & \multirow{2}{*}{SSL} & 8.03$\pm$0.45 & 7.45$\pm$1.09 & 7.23$\pm$0.17 & 6.34$\pm$2.58 & 8.23$\pm$0.07 & 8.34$\pm$1.49 & 8.12$\pm$1.26 & 10.23$\pm$3.80 & 7.34$\pm$2.06 & 7.21$\pm$1.17 & 7.43$\pm$0.22 & 7.03$\pm$1.95 & 7.75$\pm$0.27 \\
FixMatch \citep{SohnBCZZRCKL20} & & 6.37$\pm$0.52 & 7.01$\pm$0.35 & 5.28$\pm$0.10 & 6.87$\pm$0.63 & 8.40$\pm$1.69 & 9.61$\pm$1.90 & 9.33$\pm$1.72 & 9.95$\pm$0.22 & 8.34$\pm$0.48 & 7.23$\pm$0.53 & 5.67$\pm$0.74 & 5.82$\pm$0.68 & 7.49$\pm$0.27 \\
\midrule
DAN \citep{long2015learning} & \multirow{3}{*}{UDA} & 30.5$\pm$0.4 & 40.7$\pm$0.2 & 23.0$\pm$0.2 & 26.0$\pm$2.5 & 40.2$\pm$0.5 & \textbf{45.4}$\pm$2.3 & 39.0$\pm$1.3 & 40.4$\pm$0.6 & 32.2$\pm$0.5 & 32.0$\pm$2.4 & 28.2$\pm$0.2 & 27.8$\pm$1.5 & 33.8$\pm$0.4 \\
MCD \citep{saito2018maximum} & & 31.0$\pm$1.9 & 40.9$\pm$0.3 & 25.4$\pm$2.6 & 25.4$\pm$0.9 & 39.7$\pm$0.6 & 44.0$\pm$1.3 & 39.1$\pm$1.4 & 42.5$\pm$0.4 & 32.1$\pm$0.3 & \textbf{35.6}$\pm$1.3 & 27.7$\pm$0.1 & 30.3$\pm$1.9 & 34.5$\pm$0.4 \\
MDD \citep{Zhang2019BridgingTA} & & 34.6$\pm$2.2 & \textbf{41.3}$\pm$0.4 & 28.1$\pm$3.7 & 24.6$\pm$3.7 & \textbf{41.4}$\pm$3.2 & 44.4$\pm$1.5 & 38.5$\pm$3.7 & 44.2$\pm$2.4 & 32.3$\pm$0.1 & 35.1$\pm$1.8 & 29.3$\pm$0.9 & \textbf{30.6}$\pm$1.7 & 35.4$\pm$0.4 \\
SHOT \citep{Liang2020DoWR} & & 30.90$\pm$1.57 & 40.06$\pm$0.38 & 27.07$\pm$4.06 & 27.49$\pm$2.76 & 41.25$\pm$2.21 & 43.99$\pm$3.11 & 34.85$\pm$3.10 & 44.10$\pm$1.94 & 33.59$\pm$2.46 & 33.68$\pm$0.36 & 29.50$\pm$1.43 & 29.48$\pm$0.83 & 34.66$\pm$0.19 \\
\midrule
OCDA \citep{Liu2020OpenCD} & \multirow{2}{*}{MTDA} & 21.46$\pm$2.43 & 34.19$\pm$2.48 & 20.45$\pm$0.84 & 18.61$\pm$1.41 & 35.18$\pm$0.13 & 21.40$\pm$2.19 & 18.39$\pm$0.22 & 21.51$\pm$0.82 & 21.01$\pm$1.27 & 11.11$\pm$0.51 & 21.34$\pm$1.40 & 27.21$\pm$0.63 & 22.66$\pm$0.34 \\
BTDA \citep{Chen2019BlendingTargetDA} & & 19.30$\pm$3.46 & 21.36$\pm$1.45 & 24.63$\pm$1.28 & 16.86$\pm$0.84 & 20.86$\pm$0.33 & 18.02$\pm$1.30 & 16.78$\pm$0.39 & 27.81$\pm$0.47 & 19.82$\pm$0.33 & 10.22$\pm$0.66 & 11.46$\pm$5.03 & 23.36$\pm$0.80 & 19.21$\pm$0.70 \\
\midrule
GUD \citep{Volpi2018GeneralizingTU} & \multirow{4}{*}{DG} & 32.23$\pm$0.15 & 41.23$\pm$1.02 & 26.83$\pm$0.68 & 25.19$\pm$3.45 & 39.95$\pm$1.26 & 44.17$\pm$0.97 & 37.89$\pm$3.64 & 43.85$\pm$1.56 & 34.56$\pm$2.93 & 34.74$\pm$0.10 & 28.60$\pm$1.50 & 29.99$\pm$0.92 & 34.94$\pm$0.19 \\
JiGen \citep{Carlucci2019DomainGB} &  & 16.83$\pm$0.53 & 32.19$\pm$0.33 & 27.24$\pm$0.33 & 14.19$\pm$0.46 & 35.83$\pm$0.31 & 17.88$\pm$1.85 & 24.37$\pm$1.12 & 32.03$\pm$1.76 & 25.13$\pm$0.55 & 20.82$\pm$1.46 & 21.71$\pm$0.69 & 25.74$\pm$1.42 & 24.50$\pm$0.21 \\
RSC \citep{HuangWXH20} & & 14.71$\pm$1.49 & 29.81$\pm$2.63 & 22.02$\pm$0.50 & 12.00$\pm$1.31 & 33.32$\pm$0.36 & 13.94$\pm$0.98 & 21.37$\pm$4.53 & 29.48$\pm$0.62 & 21.63$\pm$0.89 & 17.74$\pm$0.34 & 18.85$\pm$1.06 & 23.13$\pm$0.83 & 21.50$\pm$0.28 \\
M-ADA \citep{qiao2020learning} & & 9.62$\pm$0.31 & 27.46$\pm$1.63 & 17.37$\pm$0.02 & 11.40$\pm$1.18 & 32.19$\pm$0.20 & 13.14$\pm$0.80 & 17.01$\pm$0.72 & 15.04$\pm$0.80 & 14.26$\pm$2.50 & 19.05$\pm$2.25 & 10.16$\pm$0.38 & 19.76$\pm$1.63 & 17.21$\pm$0.23 \\
\midrule
\textbf{DSBF} & SLDG & \textbf{35.30}$\pm$0.31 & 39.26$\pm$0.79 & \textbf{43.80}$\pm$0.31 & \textbf{33.94}$\pm$0.66 & 38.50$\pm$0.11 & 34.05$\pm$1.30 & \textbf{44.41}$\pm$0.43 & \textbf{48.89}$\pm$0.82 & \textbf{46.21}$\pm$0.15 & 33.75$\pm$0.53 & \textbf{29.73}$\pm$0.31 & 29.96$\pm$0.39 & \textbf{38.15}$\pm$0.23 \\
\bottomrule
\end{tabular}}
\end{sidewaystable*}

\begin{sidewaystable*}
\centering
\caption{Classification accuracy (\%) for the Conventional Domain Generalization (CDG) task on PACS and Office-Home datasets. The best results are emphasized in bold.}
    \label{table-cdg}
\scalebox{0.7}[0.7]{
\renewcommand\tabcolsep{8.0pt}
\begin{tabular}{l|cccc|c|cccc|c}
\toprule
\multirow{2}{*}{Methods} &
\multicolumn{5}{c|}{PACS} & \multicolumn{5}{c}{Office-Home} \\
\cline{2-11}
 & Ar & Ca & Ph & Sk & Avg. & Ar & Cl & Pr & Rw & Avg. \\
\midrule
DeepAll \citep{Carlucci2019DomainGB} & 77.85 & 74.86 & 95.73 & 67.74 & 79.05 & 52.15 & 45.86 & 70.86 & 73.15 & 60.51 \\
MMD-AAE \citep{li2018domain} & 75.2 & 72.7 & 96.0 & 64.2 & 77.0 & 56.5 & 47.3 & 72.1 & 74.8 & 62.7 \\
RSC \citep{HuangWXH20} & 83.43 & \textbf{80.31} & 95.99 & 80.85 & 85.15 & 58.42 & 47.90 & 71.63 & 74.54 & 63.12 \\
CrossGrad \citep{shankar2018generalizing} & 79.8 & 76.8 & 96.0 & 70.2 & 80.7 & 58.4 & \textbf{49.4} & 73.9 & 75.8 & 64.4 \\
D-SAMs \citep{d2018domain} & 77.33 & 72.43 & 95.30 & 77.83 & 80.72 & 58.03 & 44.37 & 69.22 & 71.45 & 60.77 \\
DSON \citep{Seo2020LearningTO} & \textbf{84.67} & 77.65 & 95.87 & \textbf{82.23} & 85.11 & 59.37 & 44.70 & 71.84 & 74.68 & 62.90 \\ 	
JiGen \citep{Carlucci2019DomainGB} & 79.42 & 75.25 & \textbf{96.03} & 71.35 & 80.51 & 53.04 & 47.51 & 71.47 & 72.79 & 61.20 \\
\textbf{DSBF} & 84.13$\pm$0.12 & 79.32$\pm$0.19 & 95.77$\pm$0.32 & 81.58$\pm$0.18 & \textbf{85.20}$\pm$0.02 
& \textbf{60.65}$\pm$0.13 & 47.33$\pm$0.04 & \textbf{74.11}$\pm$0.08 & \textbf{75.89}$\pm$0.26 & \textbf{64.50}$\pm$0.04 \\
\bottomrule
\end{tabular}}
\end{sidewaystable*}

\textbf{Implementation details.}
Following \citep{Carlucci2019DomainGB, Dou2019DomainGV, HuangWXH20}, we employ the pre-trained ResNet-18 \citep{he2016deep} as the feature extractor $b\circ{g}$ for all the experiments. The architecture of each projection networks $\{v_{j}\}_{j=2}^{K}$ is a fully-connected layer with 256 units. The classifier is a fully-connected layer with the same units as the image classes. 
For Alg. \ref{alg}, we train the model by SGD optimizer with batchsize 64, learning rate 0.01, momentum 0.9, and weight decay 0.001. 
In order to achieve efficient and stable training of Eq. (\ref{equation-fp}-\ref{equation-bf}), in each iteration, we sample data batches from 4 random classes (size of each batch is 16 for each class), we then calculate the loss within each class and obtain the final average loss to update model parameters.
The epochs for model initialization and bias filtering are both set to 20, 30, 20, 10 on PACS, Office-Home, Office-Caltech-Home, DomainNet datasets respectively. We split dataset by 0.9/0.1 for training/validation. 
Note that we report the average classification accuracy of 3 runs with different random seeds for the experiments of the SLDG task. 
We implement the baseline methods based on their source code and report the results with two decimals. The DAN, MCD, and MDD methods are implemented based on the Transfer Learning Library \url{https://github.com/thuml/Transfer-Learning-Library} which reports the results with one decinal. For the experiments of the CDG task, we cite the results of the baseline DG methods in the related papers. We directly use the groundtruth labels rather than the pseudo labels from clustering in the CDG task. 
Since all the source datasets are labeled in the CDG task, we are allowed to choose one source dataset as the $\mathcal{D}^{1}$. In our experiments on PACS dataset, when the target domain is Ar or Ph, we let Sk be the labeled source dataset $\mathcal{D}^{1}$; and when the target domain is Ca and Sk, we let Ph be the labeled source dataset $\mathcal{D}^{1}$. 
For Office-Home dataset, when the target domain is Ar or Cl, we let Pr be the labeled source dataset $\mathcal{D}^{1}$; and when the target domain is Pr or Rw, we let Cl be the labeled source dataset $\mathcal{D}^{1}$. In model initialization, we use all the multi-source data for the CDG task, and only use $\mathcal{D}^{1}$ for the SLDG task. 
We use default hyper-parameters, i.e., $\lambda$ and $\gamma$ are set to 1, in the main experiments, and further analyze their sensitivity later. 

\begin{figure*}[t]
    \centering
    \includegraphics[trim={0cm 0cm 0cm 0cm},clip,width=0.65\columnwidth]{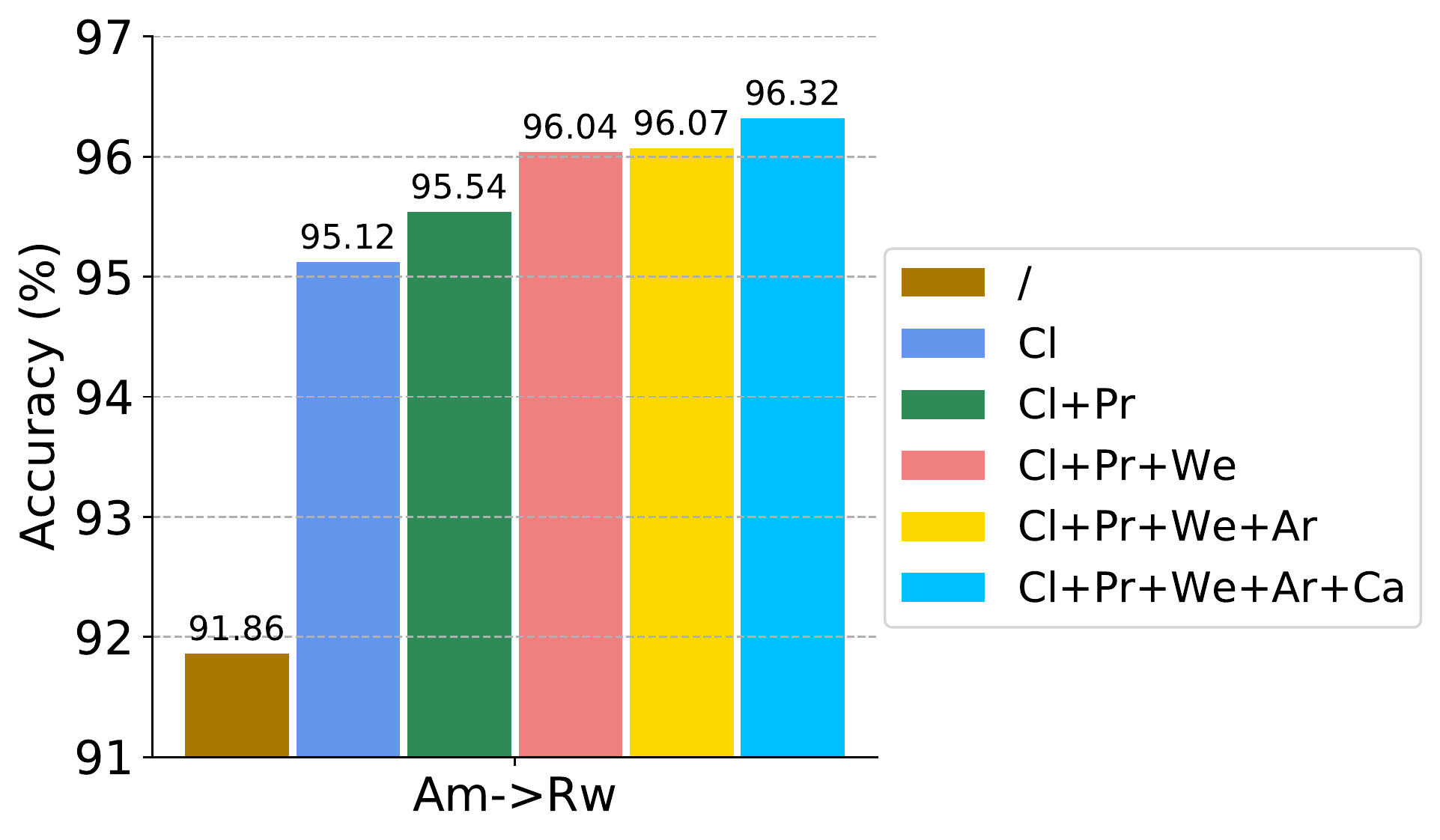}
    \includegraphics[trim={0cm 0cm 0cm 0cm},clip,width=0.65\columnwidth]{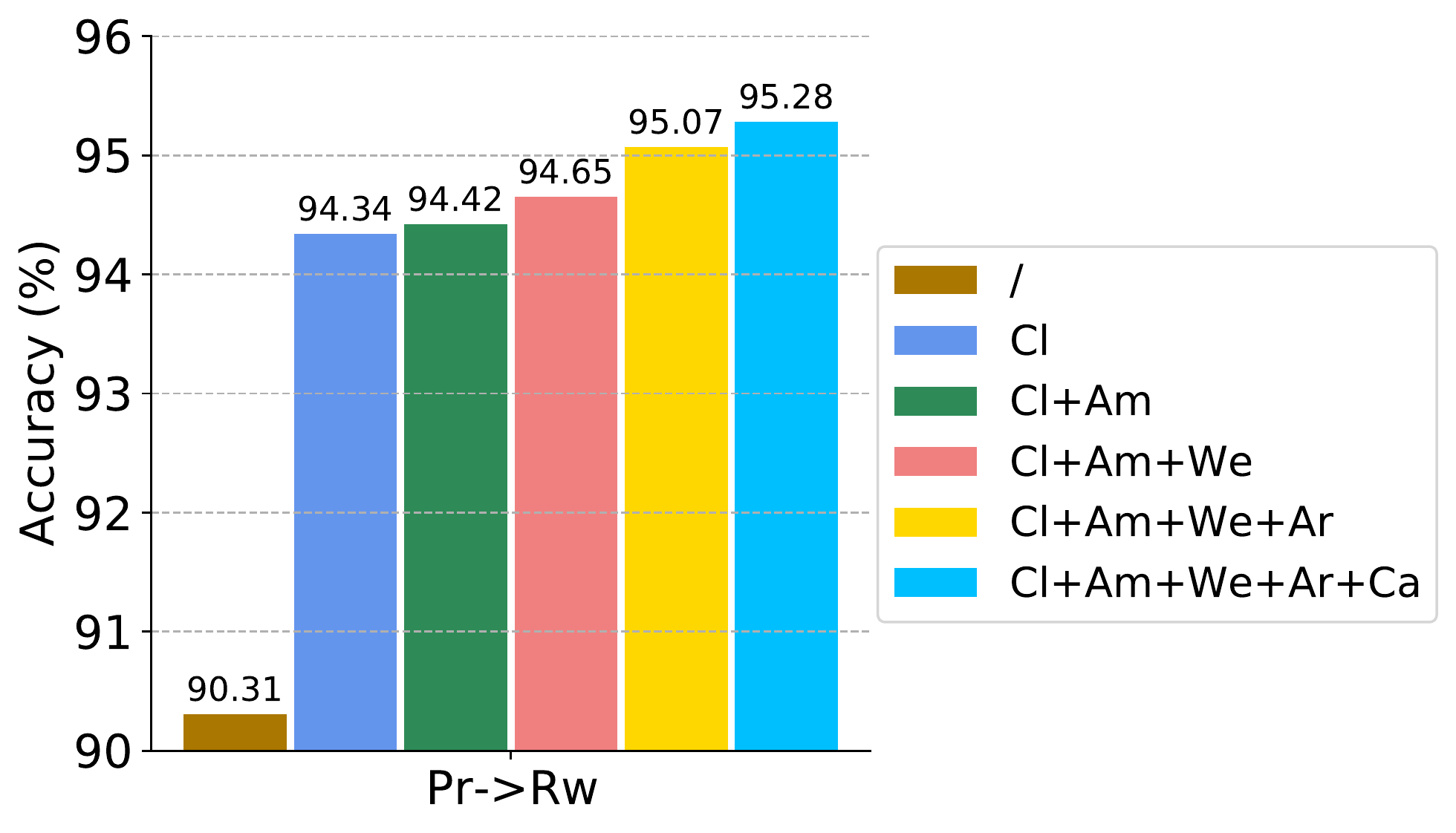}
    \includegraphics[trim={0cm 0cm 0cm 0cm},clip,width=0.65\columnwidth]{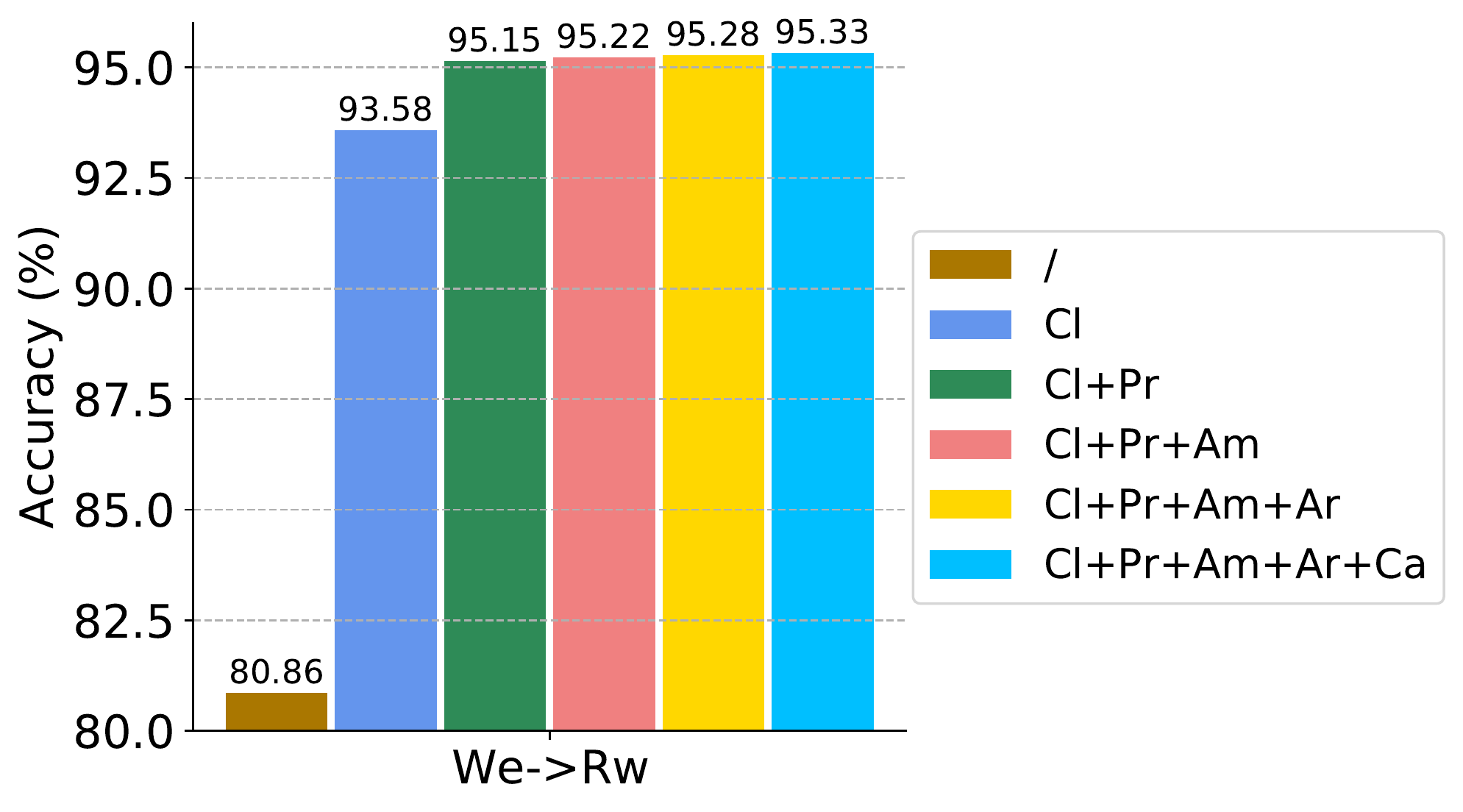}
    \caption{Accuracy for the SLDG task on Office-Caltech-Home dataset. A$\rightarrow$B represents using A, B, and other domains as the labeled source, target, and unlabeled source domains, respectively. We add one unlabeled source dataset each time from the unlabeled source domain set for each experiment. If no unlabeled source dataset is given (marked with `` / ''), the experiments are implemented in the supervised learning setting, i.e., using ERM \citep{vapnik1992principles} method.}
    \label{figure-office-caltech-home}
\end{figure*}

\begin{figure*}[t]
    \centering
    \includegraphics[trim={0cm 0cm 0cm 0cm},clip,width=1.99\columnwidth]{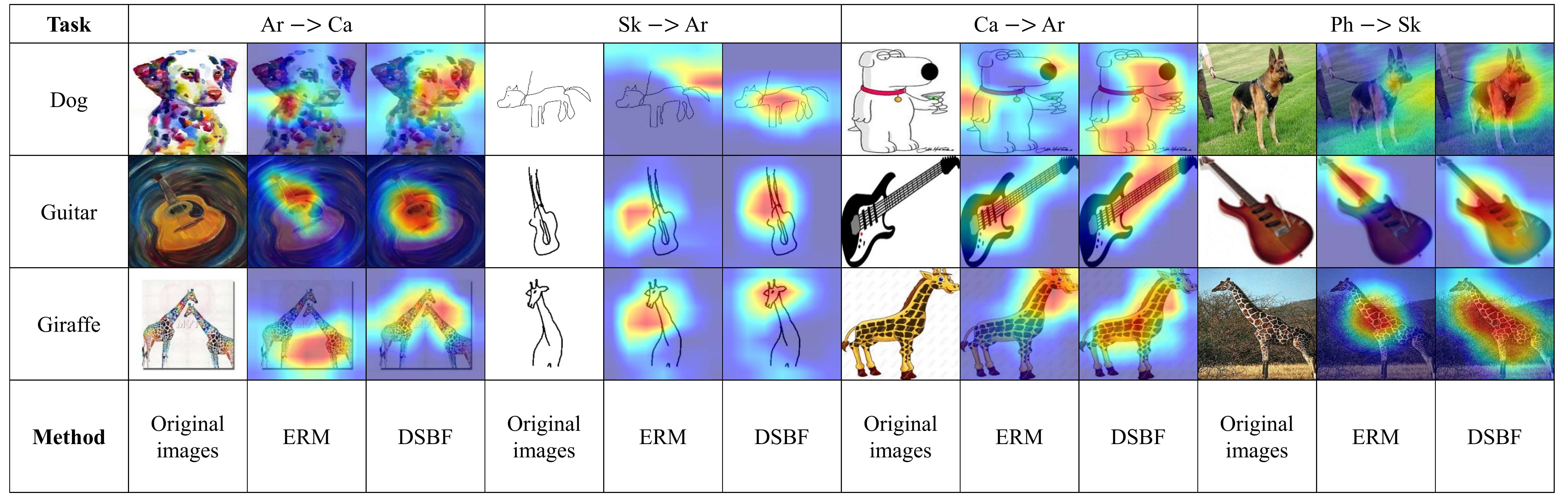}
    \caption{Grad-CAM visualization \citep{selvaraju2017grad} of the semantic information learned by the supervised learning method ERM \citep{vapnik1992principles} and the proposed method DSBF. The regions in the darker red are considered more important for the trained model to perform object recognition.}
    \label{fig-gradcam}
\end{figure*}

\begin{figure*}[t]
    \centering
    \includegraphics[trim={0cm 0cm 0cm 0cm},clip,width=0.98\columnwidth]{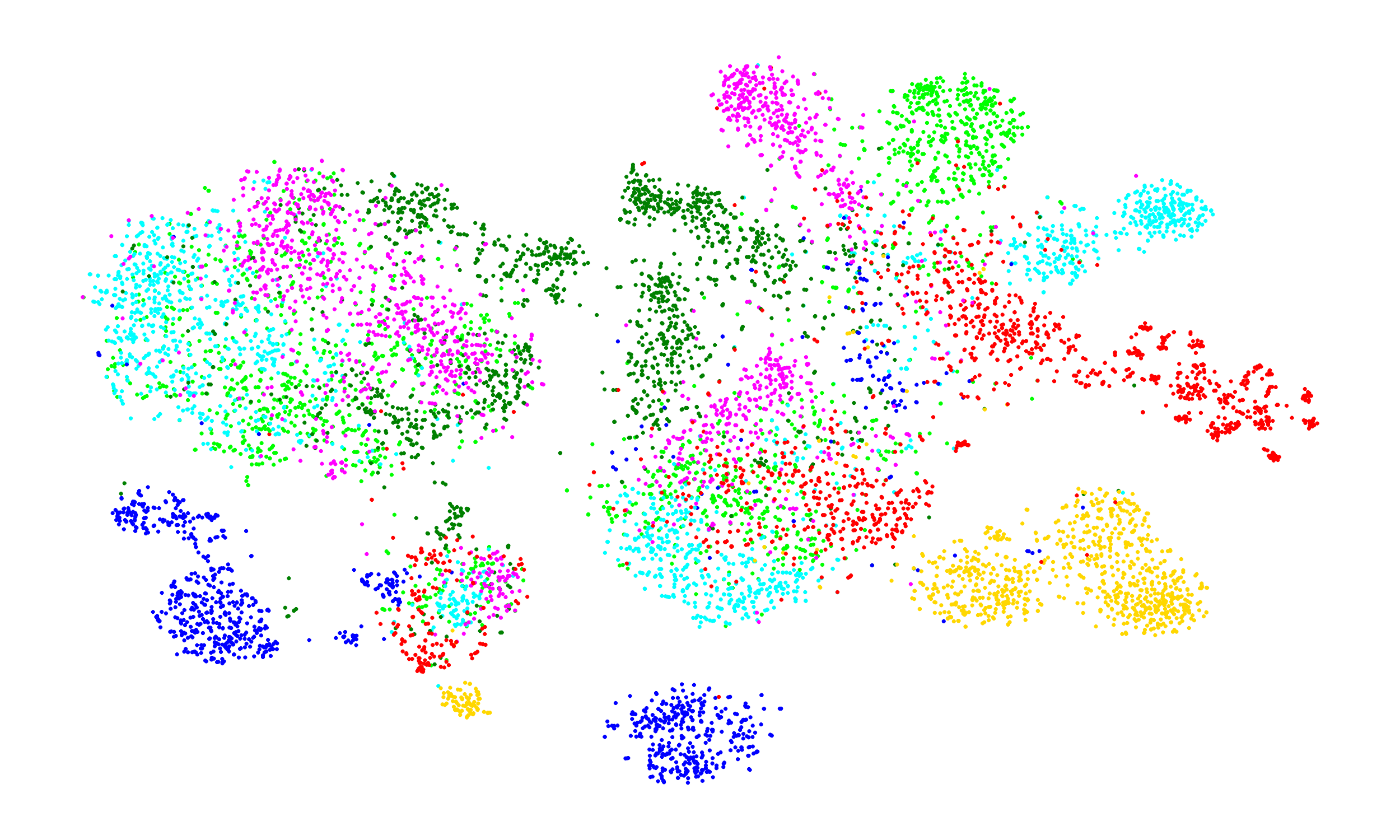}
    \includegraphics[trim={0cm 0cm 0cm 0cm},clip,width=0.98\columnwidth]{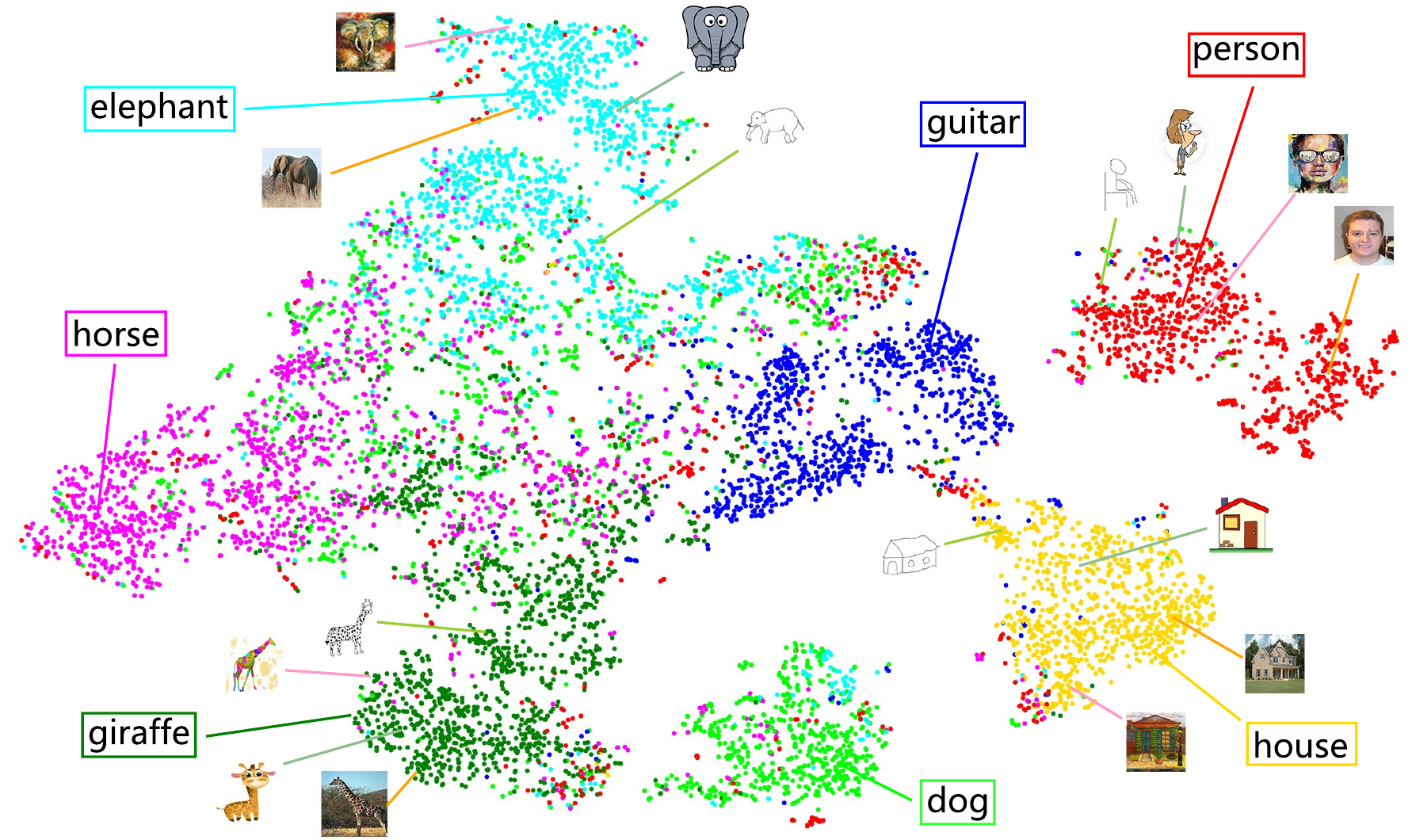}
    \caption{T-SNE visualization of the distributions of the extracted semantic feature on PACS dataset (Ph$\rightarrow$Sk), where each color represent a class. \textbf{Left}: After the model initialization stage. \textbf{Right}: After the bias filtering stage.}
    \label{fig-tsne}
\end{figure*}

\subsection{Results for the SLDG Task}
Table \ref{table-dgsls-pacs} and Table \ref{table-dgsls-home} report the results of the SLDG task on PACS and Office-Home datasets, respectively. 
We first note that the SSL methods, i.e., Mean Teacher and FixMatch, fail badly, which is probably because they rely on the i.i.d. assumption and hence severely overfit the labeled and unlabeled datasets that actually sampled from different domains/distributions. 
The next observation is that the DG methods, i.e., GUD, JiGen, RSC, M-ADA, show comparable performance to the standard SL method ERM, which is probably because they can not identify the domain invariance well by only utilizing one labeled source data. 
Since the UDA methods address the dataset shift by using both the labeled and the unlabeled data, they are allowed to learn more effective domain-invariant semantic information and hence perform obviously better. 
The reason for the worse performance of the MTDA methods, i.e. OCDA and BTDA, may be that they need some strong assumptions. For example, OCDA \citep{Liu2020OpenCD} considers a more homogeneous setting that the domain divergence is indistinct, and it directly employs the model predictions of the unlabeled data as pseudo labels for the model training. 
In comparison, the proposed DSBF method performs the best on 5 and 6 sub-tasks of the 12 sub-tasks on PACS and Office-Home datasets, respectively, and achieves the highest average accuracy which is much higher than other methods on both datasets. 
We argue that it is because DSBF method makes full use of the unlabeled source data to filter out the domain-specific bias and captures the invariant correlation between the semantic features and the labels, resulting in a well generalizable model for out-of-distribution target data. 

We then report the results of the SLDG task on a more challenging large-scale dataset, i.e., DomainNet, in Tabel \ref{table-dgsls-domainnet}. It still shows that the semi-supervised methods may fail to learn generalization from data with distribution shift. The DG methods which can only make use of the labeled source data, especially JiGen, RSC, and M-ADA, performs worse than the UDA methods and our method in this label-limited scenario. Despite the UDA methods achieve good performance, they are still surpassed by our method of DSBF since they do not prepare for the generalization on the unseen target domains. We show that DSBF still yields superior out-of-distribution generalization performance on the large-scale dataset.

To further evaluate the generalization performance gain from the unlabeled source data, we consider the scenarios with more domains using Office-Caltech-Home dataset as shown in Fig. \ref{figure-office-caltech-home}. We find that the performance improves obviously when only one unlabeled source dataset is used, especially in the third group (on the right of Fig. \ref{figure-office-caltech-home}) where the labeled source domain is We and the target domain is Rw, the utilization of the unlabeled source domain Cl significantly improves the accuracy from 80.86\% to 93.58\%. Moreover, we observe a gradual improvement in performance when given more unlabeled source domains. It indicates that DSBF only needs one unlabeled source data to perform effective domain-specific bias filtering and domain invariance learning. The bias filtering can work better when given more unlabeled source domains, which we attribute to the invariance learning of the multi-source data under the inter-domain attention mechanism.

\subsection{Results for the CDG Task}
We report the results for the CDG task on PACS and Office-Home datasets in Table \ref{table-cdg}. 
We observe that the proposed DSBF method achieves the highest average classification accuracy on both PACS and Office-Home datasets, and performs the best on more than half CDG sub-tasks on Office-Home dataset. 
DSBF method has excellent performance in effectively training a generalizable model not only in the challenging SLDG task but also in the CDG task, which illustrates the versatility of the proposed domain-specific bias filtering strategy that domain-specific bias of one domain can be filtered out by effectively employing the data of other source domains.

\begin{figure}[t]
    \centering
    \includegraphics[trim={0cm 0.5cm 0.5cm 0.5cm},clip,width=0.99\columnwidth]{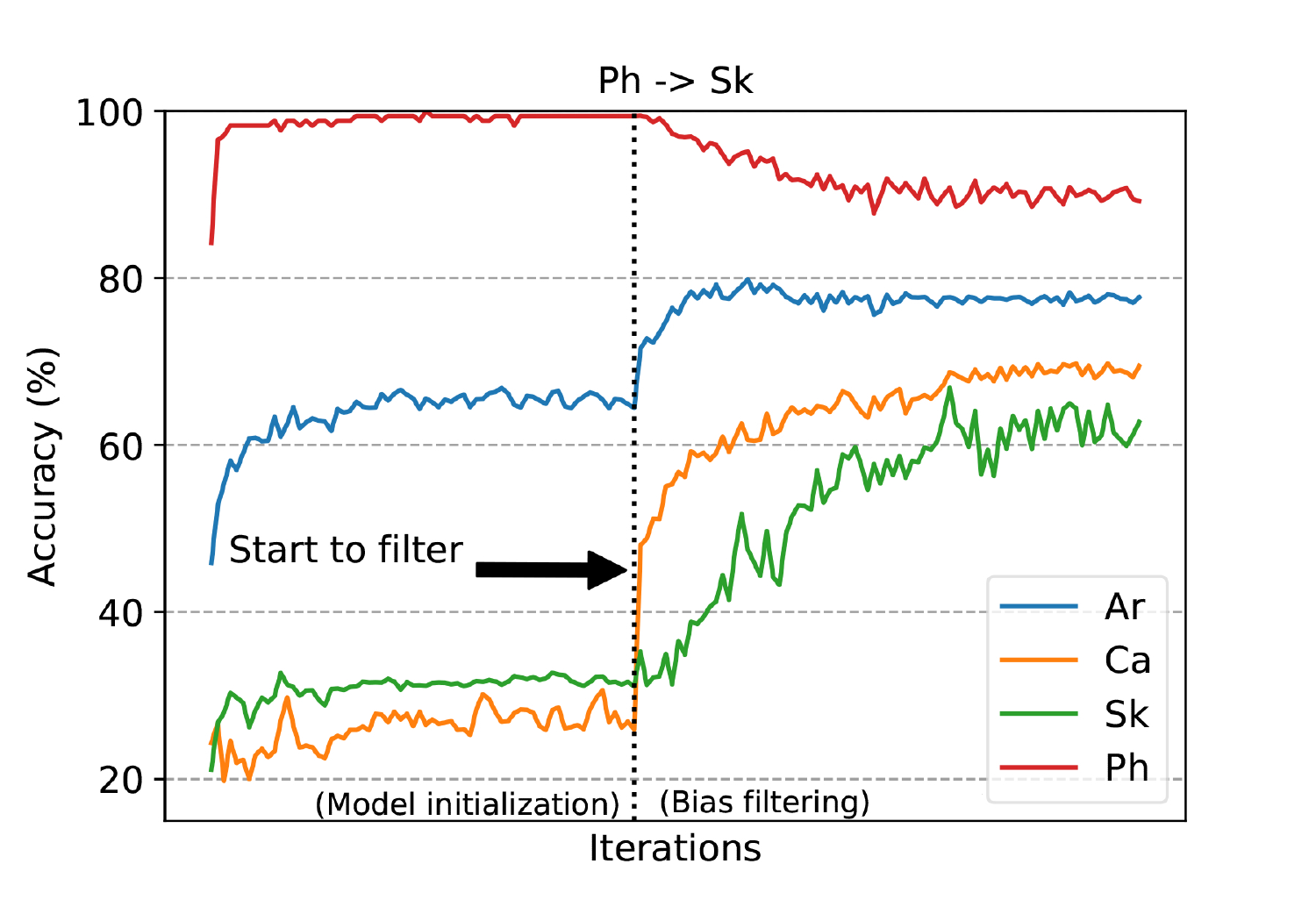}
    \caption{Classification accuracy on PACS dataset during the model initialization stage and the bias filtering stage (the labeled source domain: Ph; the target domain: Sk; the unlabeled source domains: Ar and Ca).}
    \label{fig-acc}
\end{figure}


\begin{table}
\centering
\caption{Ablation study of classification accuracy (\%) on PACS and Office-Home datasets. \textbf{DEB:} feature extractor debiasing; \textbf{REC:} classifier rectification; \textbf{ATT:} the inter-domain attention module in the classifier rectification. The best results are emphasized in bold.}
    \label{table-ablation}
\scalebox{1}[1]{
\renewcommand\tabcolsep{3.0pt}
\begin{tabular}{ccccc|cc}
\toprule
\multicolumn{2}{c}{DEB} & \multicolumn{2}{c}{REC} & \multirow{2}{*}{ATT} & 
\multirow{2}{*}{PACS} & 
\multirow{2}{*}{Home} \\
$\mathcal{L}_{IM}$ & $\mathcal{L}_{CU}$ & $\mathcal{L}_{FP}$ & $\mathcal{L}_{BF}$ & & & \\
\midrule
&  &  &  &  & 61.11 & 53.62 \\
&  & $\checkmark$ & $\checkmark$ & $\checkmark$ & 61.47 & 53.64 \\
& $\checkmark$ & $\checkmark$ & $\checkmark$ & $\checkmark$ & 65.34 & 54.34 \\
$\checkmark$ &  & $\checkmark$ & $\checkmark$ & $\checkmark$ & 65.75 & 55.11 \\
$\checkmark$ & $\checkmark$ &  &  &  & 65.99 & 55.89 \\
$\checkmark$ & $\checkmark$ & $\checkmark$ &  &  & 66.01 & 55.90 \\
$\checkmark$ & $\checkmark$ &  & $\checkmark$ &  & 66.21 & 55.92 \\
$\checkmark$ & $\checkmark$ & $\checkmark$ & $\checkmark$ &  & 66.99 & 55.97 \\
$\checkmark$ & $\checkmark$ & $\checkmark$ & $\checkmark$ & $\checkmark$ & \textbf{67.12} & \textbf{56.40} \\
\bottomrule
\end{tabular}}
\end{table}

\begin{figure}[t]
    \centering
    \includegraphics[trim={0cm 0cm 0cm 0cm},clip,width=0.46\columnwidth]{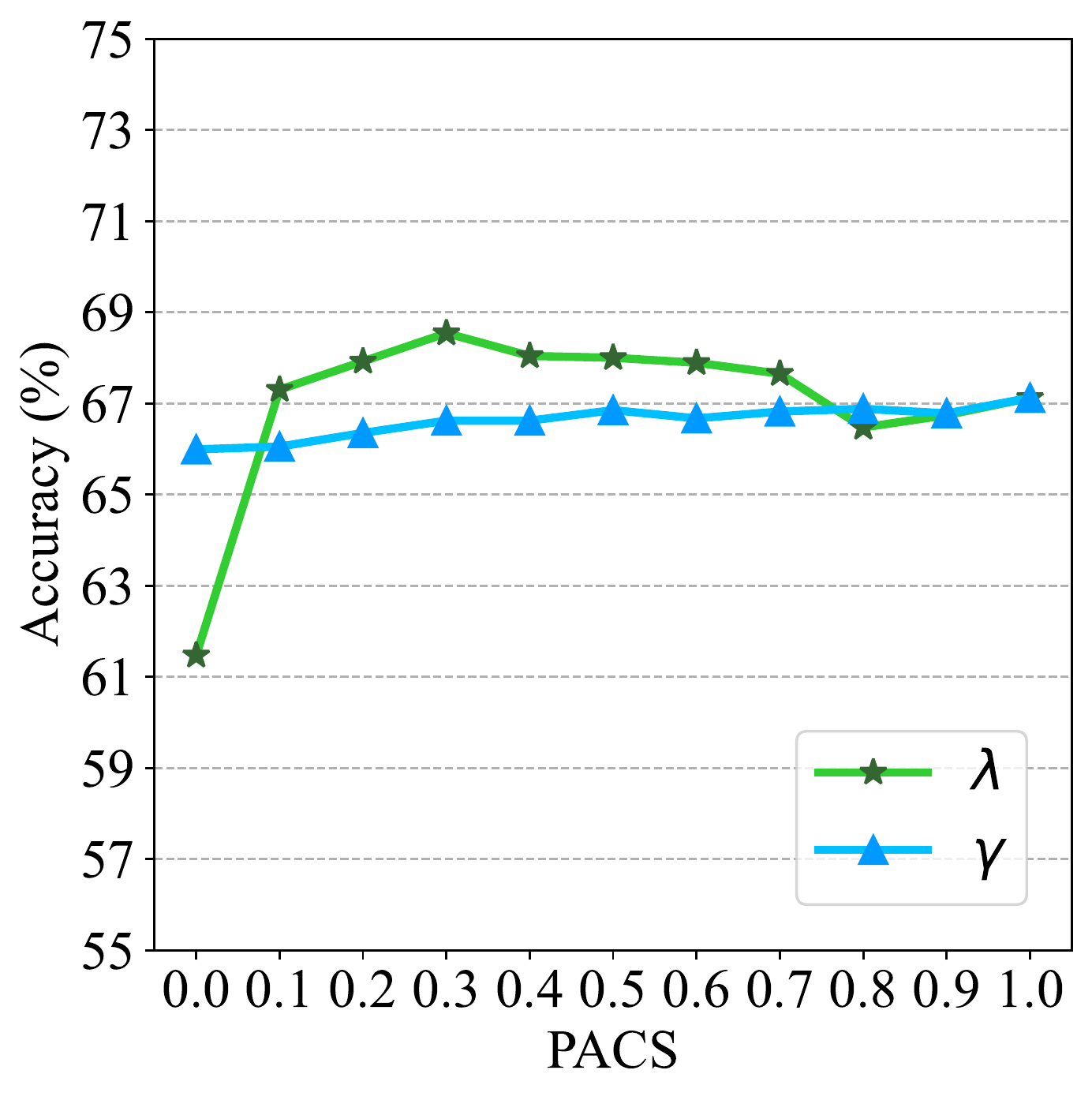}
    \includegraphics[trim={0cm 0cm 0cm 0cm},clip,width=0.46\columnwidth]{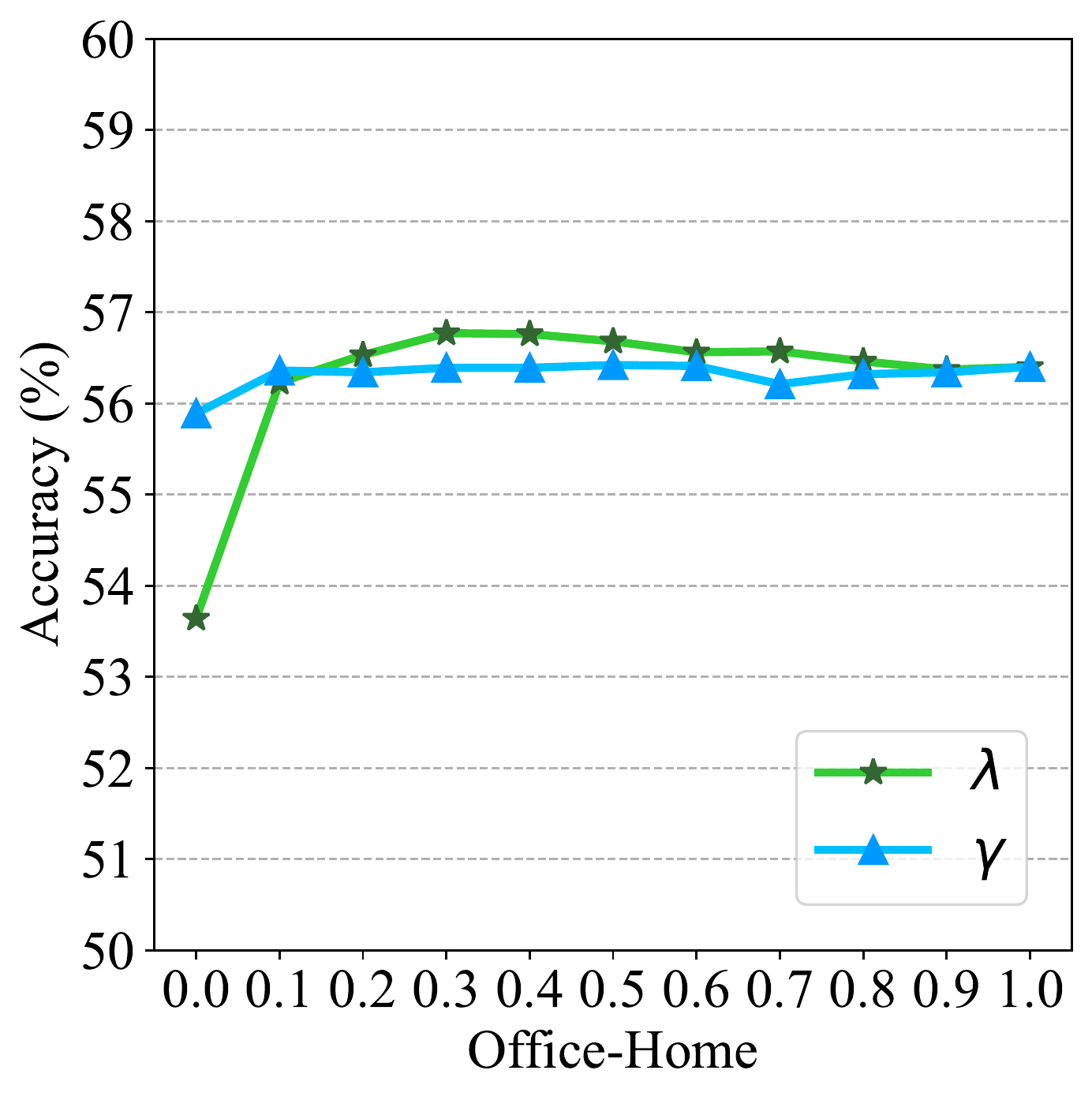}
    \caption{Sensitivity analysis of the hyper-parameters $\lambda$ and $\gamma$ for the SLDG task, which are used for the feature extractor debiasing and the classifier rectification, respectively.}
    \label{fig-hyper}
\end{figure}

\begin{figure}[t]
    \centering
    \includegraphics[trim={0cm 0cm 0cm 0cm},clip,width=0.46\columnwidth]{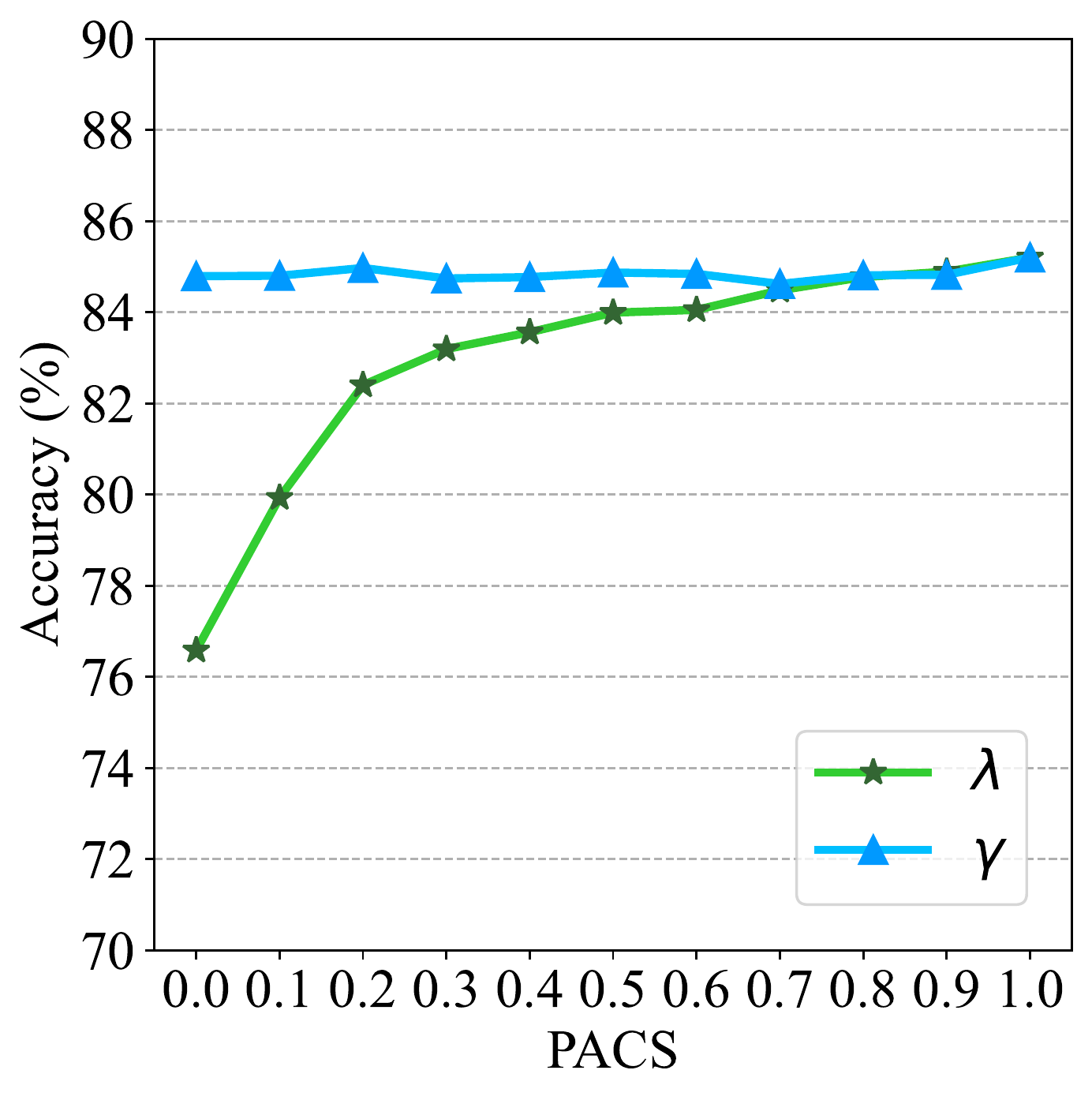}
    \includegraphics[trim={0cm 0cm 0cm 0cm},clip,width=0.46\columnwidth]{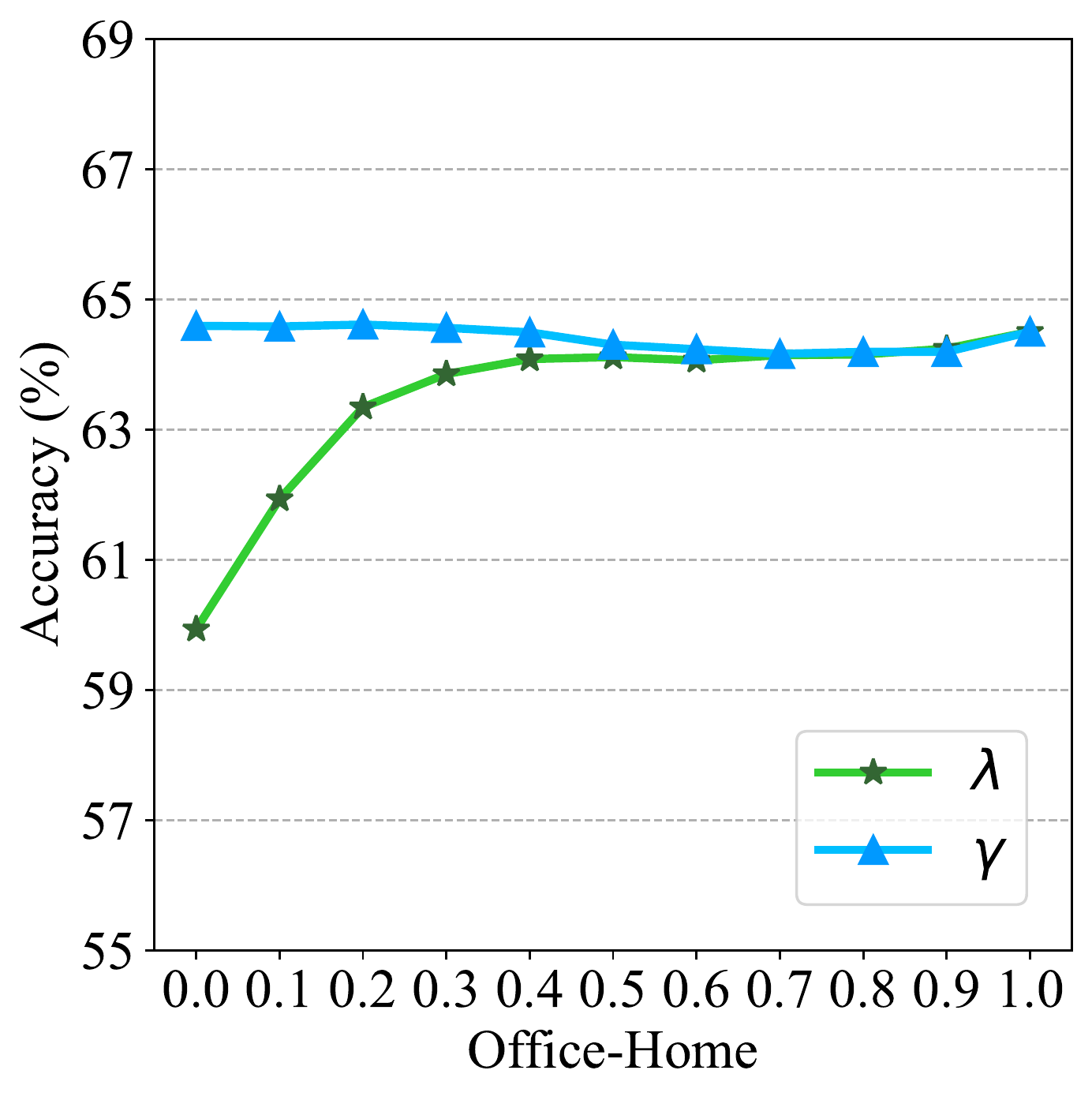}
    \caption{Sensitivity analysis of the hyper-parameters $\lambda$ and $\gamma$ for the CDG task, which are used for the feature extractor debiasing and the classifier rectification, respectively.}
    \label{fig-hyperDG}
\end{figure}

\begin{figure}[t]
    \centering
    \includegraphics[trim={0cm 0cm 0cm 0cm},clip,width=0.99\columnwidth]{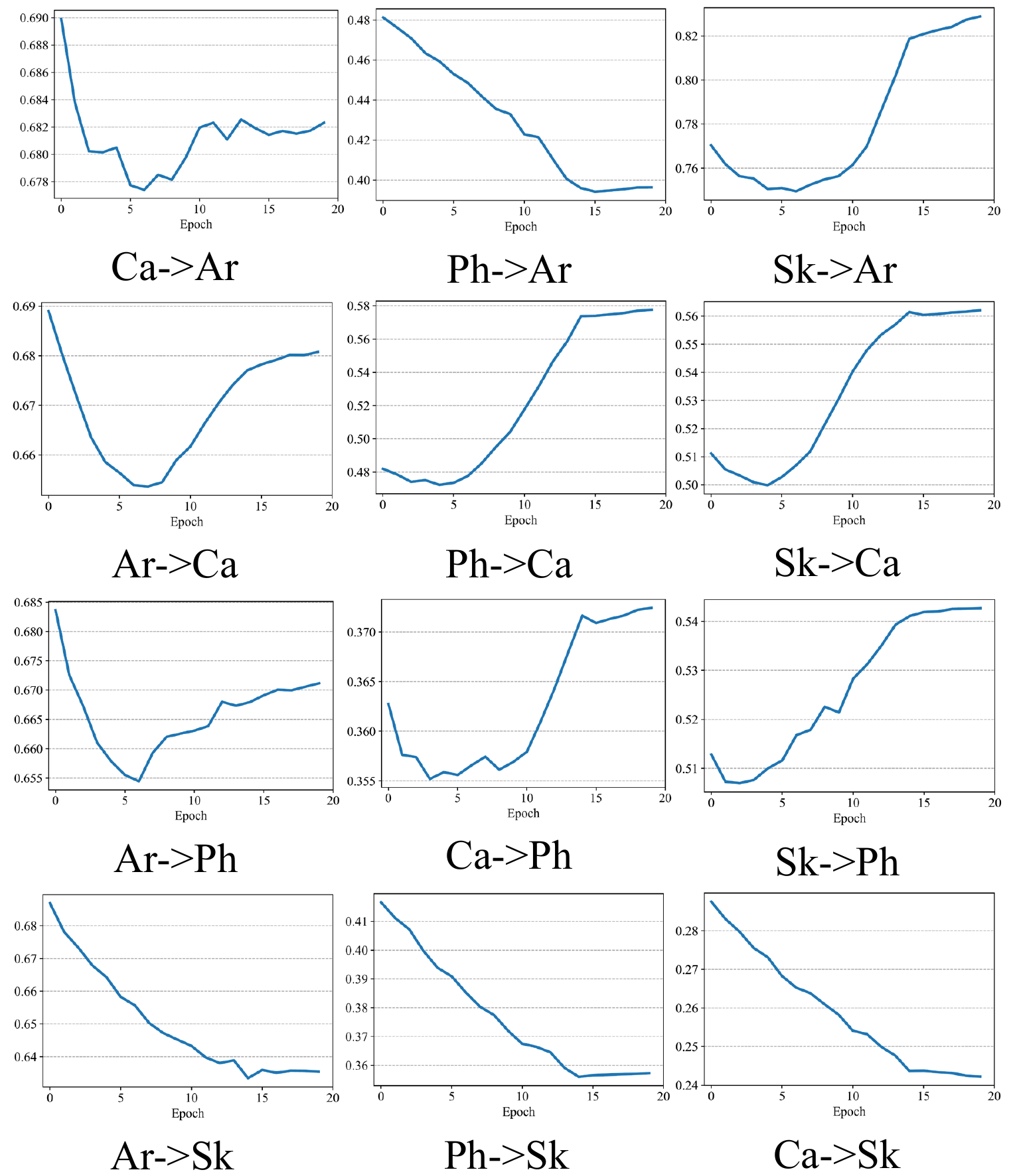}
    \caption{
    Changes of the parameter $\alpha$ of the inter-domain attention module during training on the PACS dataset.}
    \label{fig-alpha}
\end{figure}

\begin{table}
\centering
\caption{Average (\textbf{Avg.}) classification accuracy (\%) with different clustering iterations (\textbf{Iter.}) on PACS and Office-Home datasets. Iteration is 0: directly employing the model predictions as the pseudo labels. The best results are emphasized in bold.}
    \label{table-clustering}
\scalebox{1}[1]{
\renewcommand\tabcolsep{1.0pt}
\begin{tabular}{l|cccc|cccc}
\toprule
Dataset & \multicolumn{4}{c|}{PACS} & \multicolumn{4}{c}{Office-Home} \\
\midrule
Iter. & 0 & 1 & 2 & 3 & 0 & 1 & 2 & 3 \\
\midrule
Avg. & 58.28 & 67.12 & 66.87 & \textbf{67.51} & 53.30 & \textbf{56.40} & 56.16 & 56.02 \\
\bottomrule
\end{tabular}}
\end{table}

\subsection{Analysis}
\subsubsection{Semantic Invariance Learning}
Fig. \ref{fig-gradcam} shows the semantic information learned by supervised learning method ERM \citep{vapnik1992principles} (using only the labeled source data) and our method DSBF (using both the labeled and unlabeled source data). We find that DSBF employs more effective regions of the images for visual recognition, but ERM fails to pay attention to the most effective regions. It demonstrates that DSBF makes full use of the unlabeled source data to filter out the domain-specific bias in the initialized model and capture the effective semantic information for accurate object recognition.

\subsubsection{Semantic Feature Extraction}
We then exploit t-SNE \citep{van2008visualizing} to analyze the semantic feature distributions after the model initialization stage and after the bias filtering stage as shown on the left and right of Fig. \ref{fig-tsne}, respectively. 
It is evident that after bias filtering, DSBF extracts more discriminative semantic features of the data by making the same-class samples gather together. The bias filtering removes the bias in the initialized model and generates a more generalizable model.

\subsubsection{Learning Process Tracking}
We plot the learning process in Fig. \ref{fig-acc}. 
It is observed that: (1) In the model initialization stage, the trained model overfits the domain-specific bias of the labeled source data Ph (red line), and its classification accuracy rises rapidly. (2) In the bias filtering stage, the unlabeled data, i.e., Ar and Ca, are employed to filter out the domain-specific bias in both the feature extractor and classifier, the classification accuracy on the labeled source domain Ph thus drops slowly, while the performance on the unlabeled source domains Ar (blue line) and Ca (orange line) domains, as well as the unseen target domain Sk (green line), improves significantly. It clearly illustrates the learning process of DSBF, which first uses the labeled source data to initialize a discriminative model and then utilizes the unlabeled source data to filter out its bias and rectify the initialized model for improving its generalization ability.

\subsubsection{Ablation Study}
Table \ref{table-ablation} shows the ablation results, where \textbf{DEB} is feature extractor debiasing, \textbf{REC} is classifier rectification, and \textbf{ATT} is the inter-domain attention module in the classifier rectification. 
We note that all the three parts, i.e., DEB, REC, and ATT, are important for DSBF to achieve the superior performance. 
We then observe that feature extractor debiasing obviously improves the performance on both datasets. It is probably because feature extractor debiasing trains the ResNet-18 network that has much more parameters for tuning than the one fully-connected (FC) layer of the classifier trained in classifier rectification (note that no matter how many parameters the attention module has, only one FC layer of the classifier is trained and will be used for testing the final performance). The attention shows its effectiveness in domain similarities learning and generalization improvement. 
It is also observed that DSBF w/o REC w/o ATT is better than SHOT \citep{Liang2020DoWR} which uses the pseudo labels to train the classifier. It indicates that training classifier with pseudo labels could yield adverse effects in the SLDG task.

\subsubsection{Sensitivity Analysis}
We give sensitivity analysis by varying the hyper-parameters $\lambda$ and $\gamma$ for the SLDG and the CDG tasks in Fig. \ref{fig-hyper} and Fig. \ref{fig-hyperDG}, respectively. For the SLDG task, it shows that the model performance is generally stable under different hyper-parameter settings. For the CDG task, the model prefers large value of $\lambda$ but is insensitive to $\gamma$. 
We argue the reason is that the groud-truth labels are given directly in the CDG task, rather than obtained via clustering in the SLDG task. Thus, for the CDG task, the labels of the unlabeled data have higher reliability and the performance would be better when assigning larger weights, i.e., $\lambda$, to the model training with the ground-truth labels.

\subsubsection{Clustering-Based Pseudo Labels}
We further analyze the performance with different iterations of clustering-based pseudo label assignment. The results are shown in Table \ref{table-clustering}. We first observe that it is necessary to employ the clustering to obtain more accurate pseudo labels and achieve significantly better generalization performance. The second observation is that no further significant improvement can be achieved by performing more iterations. Therefore, based on this empirical experience, we may use clustering to achieve better generalization performance, but it does not need to be iterated several times. 

\subsection{Trainable Weight Parameter of Attention Module}
We show the changes of the parameter $\alpha$ (see Eq. \ref{equation-q}) of the attention module in Fig. \ref{fig-alpha}. 
Interestingly, it is observed that when given the same labeled source domain or the same target domain, the changes of $\alpha$ may show similar trend. For example, the three subfigures with the same labeled source domain of Sk, and the three subfigures with the same target domain of Sk. 
We argue that the reason for this phenomenon is that our attention module learns from the similarities among domains. When the labeled source domain or the target domain is given, the other domains may contain the similar common information for learning, which leads to the similar trend of $\alpha$.

\section{Conclusion}
In this paper, we investigate a practical task to address the real-world problem of high annotation costs for generalizable model learning, i.e., Single Labeled Domain Generalization (SLDG), where only one of the multiple source domains is labeled. To tackle this challenging task, we propose a novel framework called Domain-Specific Bias Filtering (DSBF), which unifies the exploration of the labeled and the unlabeled source data, through a model initialization stage and a bias filtering stage, enhancing discriminability and generalization of the model. 
Extensive experiments on multiple datasets show the superior performance of DSBF for the SLDG task and the CDG task. In future work, we may extend our work to the scenarios with multimodal data.

\section*{Acknowledgment}
This work was supported in part by National Key Research and Development Program of China (2021YFC3340300), National Natural Science Foundation of China (U20A20387, No. 62006207, No. 62037001), Young Elite Scientists Sponsorship Program by CAST(2021QNRC001), Project by Shanghai AI Laboratory (P22KS00111), the StarryNight Science Fund of Zhejiang University Shanghai Institute for Advanced Study (SN-ZJU-SIAS-0010), Natural Science Foundation of Zhejiang Province (LZ22F020012, LQ21F020020), Fundamental Research Funds for the Central Universities (226-2022-00142, 226-2022-00051), National Key Research and Development Project (2022YFC2504605).


\bibliography{sn-bibliography}

\begin{thebibliography}{98}
\providecommand{\natexlab}[1]{#1}
\providecommand{\url}[1]{{#1}}
\providecommand{\urlprefix}{URL }
\providecommand{\doi}[1]{\url{https://doi.org/#1}}
\providecommand{\eprint}[2][]{\url{#2}}
 \bibcommenthead

\bibitem[{Bahdanau et~al(2015)Bahdanau, Cho, and Bengio}]{BahdanauCB14}
Bahdanau D, Cho K, Bengio Y (2015) Neural machine translation by jointly
  learning to align and translate. In: International Conference on Learning
  Representations (ICLR)

\bibitem[{Balaji et~al(2018)Balaji, Sankaranarayanan, and
  Chellappa}]{balaji2018metareg}
Balaji Y, Sankaranarayanan S, Chellappa R (2018) Metareg: Towards domain
  generalization using meta-regularization. In: Advances in Neural Information
  Processing Systems (NeurIPS), pp 998--1008

\bibitem[{Bellitto et~al(2021)Bellitto, Proietto~Salanitri, Palazzo, Rundo,
  Giordano, and Spampinato}]{bellitto2021hierarchical}
Bellitto G, Proietto~Salanitri F, Palazzo S, et~al (2021) Hierarchical
  domain-adapted feature learning for video saliency prediction. International
  Journal of Computer Vision (IJCV) 129(12):3216--3232

\bibitem[{Ben-David et~al(2010)Ben-David, Blitzer, Crammer, Kulesza, Pereira,
  and Vaughan}]{ben2010theory}
Ben-David S, Blitzer J, Crammer K, et~al (2010) A theory of learning from
  different domains. Machine learning 79(1-2):151--175

\bibitem[{Blanchard et~al(2011)Blanchard, Lee, and
  Scott}]{blanchard2011generalizing}
Blanchard G, Lee G, Scott C (2011) Generalizing from several related
  classification tasks to a new unlabeled sample. Advances in Neural
  Information Processing Systems (NeurIPS) 24:2178--2186

\bibitem[{Carlucci et~al(2019)Carlucci, D'Innocente, Bucci, Caputo, and
  Tommasi}]{Carlucci2019DomainGB}
Carlucci FM, D'Innocente A, Bucci S, et~al (2019) Domain generalization by
  solving jigsaw puzzles. Proceedings of the IEEE Conference on Computer Vision
  and Pattern Recognition (CVPR) pp 2224--2233

\bibitem[{Caron et~al(2018)Caron, Bojanowski, Joulin, and
  Douze}]{caron2018deep}
Caron M, Bojanowski P, Joulin A, et~al (2018) Deep clustering for unsupervised
  learning of visual features. In: Proceedings of the European Conference on
  Computer Vision (ECCV), pp 132--149

\bibitem[{Chen et~al(2021)Chen, Wang, Li, Sakaridis, Dai, and
  Van~Gool}]{chen2021scale}
Chen Y, Wang H, Li W, et~al (2021) Scale-aware domain adaptive faster r-cnn.
  International Journal of Computer Vision (IJCV) 129(7):2223--2243

\bibitem[{Chen et~al(2019)Chen, Zhuang, Liang, and
  Lin}]{Chen2019BlendingTargetDA}
Chen Z, Zhuang J, Liang X, et~al (2019) Blending-target domain adaptation by
  adversarial meta-adaptation networks. IEEE/CVF Conference on Computer Vision
  and Pattern Recognition (CVPR) pp 2243--2252

\bibitem[{Dai et~al(2020)Dai, Sakaridis, Hecker, and
  Van~Gool}]{dai2020curriculum}
Dai D, Sakaridis C, Hecker S, et~al (2020) Curriculum model adaptation with
  synthetic and real data for semantic foggy scene understanding. International
  Journal of Computer Vision (IJCV) 128(5):1182--1204

\bibitem[{Devlin et~al(2018)Devlin, Chang, Lee, and Toutanova}]{devlin2018bert}
Devlin J, Chang MW, Lee K, et~al (2018) Bert: Pre-training of deep
  bidirectional transformers for language understanding. arXiv

\bibitem[{Ding and Fu(2017)}]{ding2017deep}
Ding Z, Fu Y (2017) Deep domain generalization with structured low-rank
  constraint. IEEE Transactions on Image Processing (TIP) 27(1):304--313

\bibitem[{Dou et~al(2019)Dou, de~Castro, Kamnitsas, and
  Glocker}]{Dou2019DomainGV}
Dou Q, de~Castro DC, Kamnitsas K, et~al (2019) Domain generalization via
  model-agnostic learning of semantic features. In: Advances in Neural
  Information Processing Systems (NeurIPS)

\bibitem[{D’Innocente and Caputo(2018)}]{d2018domain}
D’Innocente A, Caputo B (2018) Domain generalization with domain-specific
  aggregation modules. In: German Conference on Pattern Recognition, Springer,
  pp 187--198

\bibitem[{Fu et~al(2019)Fu, Liu, Tian, Li, Bao, Fang, and Lu}]{fu2019dual}
Fu J, Liu J, Tian H, et~al (2019) Dual attention network for scene
  segmentation. In: Proceedings of the IEEE Conference on Computer Vision and
  Pattern Recognition (CVPR), pp 3146--3154

\bibitem[{Ganin et~al(2016)Ganin, Ustinova, Ajakan, Germain, Larochelle,
  Laviolette, Marchand, and Lempitsky}]{ganin2016domain}
Ganin Y, Ustinova E, Ajakan H, et~al (2016) Domain-adversarial training of
  neural networks. The Journal of Machine Learning Research (JMLR)
  17(1):2096--2030

\bibitem[{Gholami et~al(2020)Gholami, Sahu, Rudovic, Bousmalis, and
  Pavlovic}]{gholami2020unsupervised}
Gholami B, Sahu P, Rudovic O, et~al (2020) Unsupervised multi-target domain
  adaptation: An information theoretic approach. IEEE Transactions on Image
  Processing (TIP) 29:3993--4002

\bibitem[{Gong et~al(2012)Gong, Shi, Sha, and Grauman}]{gong2012geodesic}
Gong B, Shi Y, Sha F, et~al (2012) Geodesic flow kernel for unsupervised domain
  adaptation. In: 2012 IEEE conference on computer vision and pattern
  recognition, IEEE, pp 2066--2073

\bibitem[{Gong et~al(2013)Gong, Grauman, and Sha}]{Gong2013ReshapingVD}
Gong B, Grauman K, Sha F (2013) Reshaping visual datasets for domain
  adaptation. In: Advances in Neural Information Processing Systems (NIPS)

\bibitem[{Gong et~al(2014)Gong, Grauman, and Sha}]{gong2014learning}
Gong B, Grauman K, Sha F (2014) Learning kernels for unsupervised domain
  adaptation with applications to visual object recognition. International
  Journal of Computer Vision (IJCV) 109(1):3--27

\bibitem[{Gong et~al(2019)Gong, Li, Chen, and Gool}]{gong2019dlow}
Gong R, Li W, Chen Y, et~al (2019) Dlow: Domain flow for adaptation and
  generalization. In: Proceedings of the IEEE/CVF Conference on Computer Vision
  and Pattern Recognition (CVPR), pp 2477--2486

\bibitem[{He et~al(2016)He, Zhang, Ren, and Sun}]{he2016deep}
He K, Zhang X, Ren S, et~al (2016) Deep residual learning for image
  recognition. In: Proceedings of the IEEE Conference on Computer Vision and
  Pattern Recognition (CVPR), pp 770--778

\bibitem[{Hinton et~al(2015)Hinton, Vinyals, and Dean}]{hinton2015distilling}
Hinton G, Vinyals O, Dean J (2015) Distilling the knowledge in a neural
  network. arXiv preprint arXiv:150302531

\bibitem[{Ho and Gopalan(2014)}]{ho2014model}
Ho HT, Gopalan R (2014) Model-driven domain adaptation on product manifolds for
  unconstrained face recognition. International journal of computer vision
  (IJCV) 109(1-2):110--125

\bibitem[{Hoffman et~al(2012)Hoffman, Kulis, Darrell, and
  Saenko}]{hoffman2012discovering}
Hoffman J, Kulis B, Darrell T, et~al (2012) Discovering latent domains for
  multisource domain adaptation. In: European Conference on Computer Vision
  (ECCV), Springer, pp 702--715

\bibitem[{Hoffman et~al(2014)Hoffman, Rodner, Donahue, Kulis, and
  Saenko}]{hoffman2014asymmetric}
Hoffman J, Rodner E, Donahue J, et~al (2014) Asymmetric and category invariant
  feature transformations for domain adaptation. International journal of
  computer vision (IJCV) 109(1-2):28--41

\bibitem[{Huang et~al(2021)Huang, Wu, Xu, Zhong, and
  Zhang}]{huang2021unsupervised}
Huang Y, Wu Q, Xu J, et~al (2021) Unsupervised domain adaptation with
  background shift mitigating for person re-identification. International
  Journal of Computer Vision (IJCV) 129(7):2244--2263

\bibitem[{Huang et~al(2020)Huang, Wang, Xing, and Huang}]{HuangWXH20}
Huang Z, Wang H, Xing EP, et~al (2020) Self-challenging improves cross-domain
  generalization. In: Proceedings of the European Conference on Computer Vision
  (ECCV), pp 124--140

\bibitem[{Kan et~al(2014)Kan, Wu, Shan, and Chen}]{kan2014domain}
Kan M, Wu J, Shan S, et~al (2014) Domain adaptation for face recognition:
  Targetize source domain bridged by common subspace. International journal of
  computer vision (IJCV) 109(1-2):94--109

\bibitem[{Kang et~al(2019)Kang, Jiang, Yang, and
  Hauptmann}]{Kang2019ContrastiveAN}
Kang G, Jiang L, Yang Y, et~al (2019) Contrastive adaptation network for
  unsupervised domain adaptation. In: Proceedings of the IEEE Conference on
  Computer Vision and Pattern Recognition (CVPR), pp 4888--4897

\bibitem[{Kundu et~al(2020)Kundu, Venkat, Babu et~al}]{kundu2020universal}
Kundu JN, Venkat N, Babu RV, et~al (2020) Universal source-free domain
  adaptation. In: Proceedings of the IEEE Conference on Computer Vision and
  Pattern Recognition (CVPR), pp 4544--4553

\bibitem[{{LeCun} et~al(2015){LeCun}, {Bengio}, and {Hinton}}]{lecun2015deep}
{LeCun} Y, {Bengio} Y, {Hinton} G (2015) Deep learning. Nature
  521(7553):436--444

\bibitem[{Li et~al(2017)Li, Yang, Song, and Hospedales}]{li2017deeper}
Li D, Yang Y, Song YZ, et~al (2017) Deeper, broader and artier domain
  generalization. In: Proceedings of the IEEE International Conference on
  Computer Vision (ICCV), pp 5542--5550

\bibitem[{Li et~al(2019)Li, Zhang, Yang, Liu, Song, and
  Hospedales}]{Li2019EpisodicTF}
Li D, Zhang J, Yang Y, et~al (2019) Episodic training for domain
  generalization. Proceedings of the IEEE International Conference on Computer
  Vision (ICCV) pp 1446--1455

\bibitem[{Li et~al(2018)Li, Pan, Wang, and Kot}]{li2018domain}
Li H, Pan SJ, Wang S, et~al (2018) Domain generalization with adversarial
  feature learning. In: Proceedings of the IEEE Conference on Computer Vision
  and Pattern Recognition (CVPR), pp 5400--5409

\bibitem[{Li et~al(2020{\natexlab{a}})Li, Wang, Wan, Wang, Li, and
  Kot}]{Li2020DomainGF}
Li H, Wang Y, Wan R, et~al (2020{\natexlab{a}}) Domain generalization for
  medical imaging classification with linear-dependency regularization. In:
  Advances in Neural Information Processing Systems (NeurIPS)

\bibitem[{Li et~al(2021)Li, Wan, Wang, and Kot}]{li2021unsupervised}
Li H, Wan R, Wang S, et~al (2021) Unsupervised domain adaptation in the wild
  via disentangling representation learning. International Journal of Computer
  Vision (IJCV) 129(2):267--283

\bibitem[{Li et~al(2020{\natexlab{b}})Li, Cao, Wu, and Wong}]{li2020generating}
Li R, Cao W, Wu S, et~al (2020{\natexlab{b}}) Generating target image-label
  pairs for unsupervised domain adaptation. IEEE Transactions on Image
  Processing (TIP) 29:7997--8011

\bibitem[{Li et~al(2020{\natexlab{c}})Li, Hu, Li, Dong, Zhang, and
  Tian}]{li2020aligning}
Li Y, Hu W, Li H, et~al (2020{\natexlab{c}}) Aligning discriminative and
  representative features: An unsupervised domain adaptation method for
  building damage assessment. IEEE Transactions on Image Processing (TIP)
  29:6110--6122

\bibitem[{Liang et~al(2020)Liang, Hu, and Feng}]{Liang2020DoWR}
Liang J, Hu D, Feng J (2020) Do we really need to access the source data?
  source hypothesis transfer for unsupervised domain adaptation. In:
  International Conference on Machine Learning (ICML), PMLR

\bibitem[{Lin et~al(2020)Lin, Li, and Kot}]{lin2020multi}
Lin S, Li CT, Kot AC (2020) Multi-domain adversarial feature generalization for
  person re-identification. IEEE Transactions on Image Processing (TIP)
  30:1596--1607

\bibitem[{Liu et~al(2020)Liu, Miao, Pan, Zhan, Lin, Yu, and
  Gong}]{Liu2020OpenCD}
Liu Z, Miao Z, Pan X, et~al (2020) Open compound domain adaptation. IEEE/CVF
  Conference on Computer Vision and Pattern Recognition (CVPR) pp
  12,403--12,412

\bibitem[{Long et~al(2015)Long, Cao, Wang, and Jordan}]{long2015learning}
Long M, Cao Y, Wang J, et~al (2015) Learning transferable features with deep
  adaptation networks. In: International conference on machine learning (ICML),
  PMLR, pp 97--105

\bibitem[{Long et~al(2017)Long, Zhu, Wang, and Jordan}]{long2017deep}
Long M, Zhu H, Wang J, et~al (2017) Deep transfer learning with joint
  adaptation networks. In: International conference on machine learning (ICML),
  PMLR, pp 2208--2217

\bibitem[{Long et~al(2018)Long, Cao, Wang, and Jordan}]{long2018conditional}
Long M, Cao Z, Wang J, et~al (2018) Conditional adversarial domain adaptation.
  In: Advances in Neural Information Processing Systems (NeurIPS), pp
  1640--1650

\bibitem[{Van~der Maaten and Hinton(2008)}]{van2008visualizing}
Van~der Maaten L, Hinton G (2008) Visualizing data using t-sne. Journal of
  Machine Learning Research (JMLR) 9(11)

\bibitem[{Mancini et~al(2019{\natexlab{a}})Mancini, Bulo, Caputo, and
  Ricci}]{mancini2019adagraph}
Mancini M, Bulo SR, Caputo B, et~al (2019{\natexlab{a}}) Adagraph: Unifying
  predictive and continuous domain adaptation through graphs. In: Proceedings
  of the IEEE/CVF Conference on Computer Vision and Pattern Recognition (CVPR),
  pp 6568--6577

\bibitem[{Mancini et~al(2019{\natexlab{b}})Mancini, Porzi, Bulo, Caputo, and
  Ricci}]{mancini2019inferring}
Mancini M, Porzi L, Bulo SR, et~al (2019{\natexlab{b}}) Inferring latent
  domains for unsupervised deep domain adaptation. IEEE transactions on pattern
  analysis and machine intelligence (TPAMI)

\bibitem[{Matsuura and Harada(2020)}]{Matsuura2020DomainGU}
Matsuura T, Harada T (2020) Domain generalization using a mixture of multiple
  latent domains. In: Proceedings of the AAAI Conference on Artificial
  Intelligence (AAAI)

\bibitem[{Miao et~al(2022)Miao, Yuan, and Kuang}]{Miao2022DomainGV}
Miao Q, Yuan J, Kuang K (2022) Domain generalization via contrastive causal
  learning. ArXiv abs/2210.02655

\bibitem[{Peng et~al(2019)Peng, Bai, Xia, Huang, Saenko, and
  Wang}]{peng2019moment}
Peng X, Bai Q, Xia X, et~al (2019) Moment matching for multi-source domain
  adaptation. In: Proceedings of the IEEE International Conference on Computer
  Vision (ICCV), pp 1406--1415

\bibitem[{Qian et~al(2022)Qian, Xu, Lv, Zhang, Jiang, Liu, Zeng, Chua, and
  Wu}]{DBLP:conf/kdd/QianXLZJLZC022}
Qian X, Xu Y, Lv F, et~al (2022) Intelligent request strategy design in
  recommender system. In: {KDD} '22: The 28th {ACM} {SIGKDD} Conference on
  Knowledge Discovery and Data Mining. {ACM}, pp 3772--3782

\bibitem[{Qiao et~al(2020)Qiao, Zhao, and Peng}]{qiao2020learning}
Qiao F, Zhao L, Peng X (2020) Learning to learn single domain generalization.
  In: Proceedings of the IEEE Conference on Computer Vision and Pattern
  Recognition, pp 12,556--12,565

\bibitem[{Quionero-Candela et~al(2009)Quionero-Candela, Sugiyama, Schwaighofer,
  and Lawrence}]{quionero2009dataset}
Quionero-Candela J, Sugiyama M, Schwaighofer A, et~al (2009) Dataset shift in
  machine learning. The MIT Press

\bibitem[{Saito et~al(2018)Saito, Watanabe, Ushiku, and
  Harada}]{saito2018maximum}
Saito K, Watanabe K, Ushiku Y, et~al (2018) Maximum classifier discrepancy for
  unsupervised domain adaptation. In: Proceedings of the IEEE Conference on
  Computer Vision and Pattern Recognition (CVPR), pp 3723--3732

\bibitem[{Schmidhuber(1987)}]{schmidhuber1987evolutionary}
Schmidhuber J (1987) Evolutionary principles in self-referential learning, or
  on learning how to learn: the meta-meta-... hook. PhD thesis, Technische
  Universit{\"a}t M{\"u}nchen

\bibitem[{Selvaraju et~al(2017)Selvaraju, Cogswell, Das, Vedantam, Parikh, and
  Batra}]{selvaraju2017grad}
Selvaraju RR, Cogswell M, Das A, et~al (2017) Grad-cam: Visual explanations
  from deep networks via gradient-based localization. In: Proceedings of the
  IEEE international conference on computer vision (ICCV), pp 618--626

\bibitem[{Seo et~al(2020)Seo, Suh, Kim, Han, and Han}]{Seo2020LearningTO}
Seo S, Suh Y, Kim D, et~al (2020) Learning to optimize domain specific
  normalization for domain generalization. In: European Conference on Computer
  Vision (ECCV)

\bibitem[{Shankar et~al(2018)Shankar, Piratla, Chakrabarti, Chaudhuri, Jyothi,
  and Sarawagi}]{shankar2018generalizing}
Shankar S, Piratla V, Chakrabarti S, et~al (2018) Generalizing across domains
  via cross-gradient training. International Conference on Learning
  Representation (ICLR)

\bibitem[{Shen et~al(2021)Shen, Huang, Shi, Liu, Maheshwari, Zheng, Xue,
  Savvides, and Huang}]{shen2021cdtd}
Shen Z, Huang M, Shi J, et~al (2021) Cdtd: A large-scale cross-domain benchmark
  for instance-level image-to-image translation and domain adaptive object
  detection. International Journal of Computer Vision (IJCV) 129(3):761--780

\bibitem[{Sindagi and Srivastava(2017)}]{sindagi2017domain}
Sindagi VA, Srivastava S (2017) Domain adaptation for automatic oled panel
  defect detection using adaptive support vector data description.
  International Journal of Computer Vision (IJCV) 122(2):193--211

\bibitem[{Sohn et~al(2020)Sohn, Berthelot, Carlini, Zhang, Zhang, Raffel,
  Cubuk, Kurakin, and Li}]{SohnBCZZRCKL20}
Sohn K, Berthelot D, Carlini N, et~al (2020) Fixmatch: Simplifying
  semi-supervised learning with consistency and confidence. In: Advances in
  Neural Information Processing Systems (NeurIPS)

\bibitem[{Tarvainen and Valpola(2017)}]{Tarvainen2017MeanTA}
Tarvainen A, Valpola H (2017) Mean teachers are better role models:
  Weight-averaged consistency targets improve semi-supervised deep learning
  results. In: Advances in Neural Information Processing Systems (NeurIPS)

\bibitem[{Vapnik(1992)}]{vapnik1992principles}
Vapnik V (1992) Principles of risk minimization for learning theory. In:
  Advances in Neural Information Processing Systems (NeurIPS), pp 831--838

\bibitem[{Vaswani et~al(2017)Vaswani, Shazeer, Parmar, Uszkoreit, Jones, Gomez,
  Kaiser, and Polosukhin}]{VaswaniSPUJGKP17}
Vaswani A, Shazeer N, Parmar N, et~al (2017) Attention is all you need. In:
  Advances in Neural Information Processing Systems (NeurIPS), pp 5998--6008

\bibitem[{Venkateswara et~al(2017)Venkateswara, Eusebio, Chakraborty, and
  Panchanathan}]{venkateswara2017deep}
Venkateswara H, Eusebio J, Chakraborty S, et~al (2017) Deep hashing network for
  unsupervised domain adaptation. In: Proceedings of the IEEE Conference on
  Computer Vision and Pattern Recognition (CVPR), pp 5018--5027

\bibitem[{Volpi et~al(2018)Volpi, Namkoong, Sener, Duchi, Murino, and
  Savarese}]{Volpi2018GeneralizingTU}
Volpi R, Namkoong H, Sener O, et~al (2018) Generalizing to unseen domains via
  adversarial data augmentation. Advances in Neural Information Processing
  Systems (NeurIPS)

\bibitem[{Wang et~al(2017)Wang, Jiang, Qian, Yang, Li, Zhang, Wang, and
  Tang}]{wang2017residual}
Wang F, Jiang M, Qian C, et~al (2017) Residual attention network for image
  classification. In: Proceedings of the IEEE Conference on Computer Vision and
  Pattern Recognition (CVPR), pp 3156--3164

\bibitem[{Wang et~al(2021)Wang, Cheng, and Jiang}]{wang2021domain}
Wang J, Cheng MM, Jiang J (2021) Domain shift preservation for zero-shot domain
  adaptation. IEEE Transactions on Image Processing (TIP)

\bibitem[{Wang et~al(2020{\natexlab{a}})Wang, Yu, Li, Fu, and
  Heng}]{Wang2020LearningFE}
Wang S, Yu L, Li C, et~al (2020{\natexlab{a}}) Learning from extrinsic and
  intrinsic supervisions for domain generalization. In: Proceedings of the
  European Conference on Computer Vision (ECCV)

\bibitem[{Wang et~al(2020{\natexlab{b}})Wang, Kihara, Luo, and
  Qi}]{wang2020enaet}
Wang X, Kihara D, Luo J, et~al (2020{\natexlab{b}}) Enaet: A self-trained
  framework for semi-supervised and supervised learning with ensemble
  transformations. IEEE Transactions on Image Processing (TIP) 30:1639--1647

\bibitem[{Wang et~al(2020{\natexlab{c}})Wang, Zhang, Hao, and
  Song}]{wang2020attention}
Wang Y, Zhang Z, Hao W, et~al (2020{\natexlab{c}}) Attention guided multiple
  source and target domain adaptation. IEEE Transactions on Image Processing
  (TIP) 30:892--906

\bibitem[{Wu et~al(2019)Wu, Wang, Gonzalez, Goldstein, and Davis}]{wu2019ace}
Wu Z, Wang X, Gonzalez JE, et~al (2019) Ace: Adapting to changing environments
  for semantic segmentation. In: Proceedings of the IEEE/CVF International
  Conference on Computer Vision (ICCV), pp 2121--2130

\bibitem[{Xiong et~al(2014)Xiong, McCloskey, Hsieh, and
  Corso}]{xiong2014latent}
Xiong C, McCloskey S, Hsieh SH, et~al (2014) Latent domains modeling for visual
  domain adaptation. In: Proceedings of the AAAI Conference on Artificial
  Intelligence (AAAI)

\bibitem[{Xu et~al(2021)Xu, Yang, Deng, Qian, and Wang}]{xu2021neutral}
Xu H, Yang M, Deng L, et~al (2021) Neutral cross-entropy loss based
  unsupervised domain adaptation for semantic segmentation. IEEE Transactions
  on Image Processing (TIP) 30:4516--4525

\bibitem[{Xu et~al(2016)Xu, Ramos, V{\'a}zquez, and
  L{\'o}pez}]{xu2016hierarchical}
Xu J, Ramos S, V{\'a}zquez D, et~al (2016) Hierarchical adaptive structural svm
  for domain adaptation. International Journal of Computer Vision (IJCV)
  119(2):159--178

\bibitem[{Yamada et~al(2014)Yamada, Sigal, and Chang}]{yamada2014domain}
Yamada M, Sigal L, Chang Y (2014) Domain adaptation for structured regression.
  International journal of computer vision (IJCV) 109(1-2):126--145

\bibitem[{Yang et~al(2021)Yang, Song, King, and Xu}]{yang2021survey}
Yang X, Song Z, King I, et~al (2021) A survey on deep semi-supervised learning.
  arXiv preprint arXiv:210300550

\bibitem[{Yasarla et~al(2021)Yasarla, Sindagi, and Patel}]{yasarla2021semi}
Yasarla R, Sindagi VA, Patel VM (2021) Semi-supervised image deraining using
  gaussian processes. IEEE Transactions on Image Processing (TIP)

\bibitem[{Yu et~al(2018)Yu, Hu, and Chen}]{Yu2018MultitargetUD}
Yu H, Hu M, Chen S (2018) Multi-target unsupervised domain adaptation without
  exactly shared categories. arXiv abs/1809.00852

\bibitem[{Yuan et~al(2021{\natexlab{a}})Yuan, Ma, Chen, Kuang, Wu, and
  Lin}]{yuan2021collaborative}
Yuan J, Ma X, Chen D, et~al (2021{\natexlab{a}}) Collaborative semantic
  aggregation and calibration for separated domain generalization. arXiv
  e-prints pp arXiv--2110

\bibitem[{Yuan et~al(2021{\natexlab{b}})Yuan, Ma, Kuang, Xiong, Gong, and
  Lin}]{yuan2021learning}
Yuan J, Ma X, Kuang K, et~al (2021{\natexlab{b}}) Learning domain-invariant
  relationship with instrumental variable for domain generalization. arXiv
  preprint arXiv:211001438

\bibitem[{Yuan et~al(2022)Yuan, Ma, Chen, Kuang, Wu, and
  Lin}]{Yuan2022LabelEfficientDG}
Yuan J, Ma X, Chen D, et~al (2022) Label-efficient domain generalization via
  collaborative exploration and generalization. Proceedings of the 30th ACM
  International Conference on Multimedia

\bibitem[{Zhang et~al(2020{\natexlab{a}})Zhang, Zhang, and
  Li}]{zhang2020acausal}
Zhang C, Zhang K, Li Y (2020{\natexlab{a}}) A causal view on robustness of
  neural networks. In: Advances in Neural Information Processing Systems
  (NeurIPS)

\bibitem[{Zhang et~al(2019{\natexlab{a}})Zhang, Goodfellow, Metaxas, and
  Odena}]{zhang2019self}
Zhang H, Goodfellow I, Metaxas D, et~al (2019{\natexlab{a}}) Self-attention
  generative adversarial networks. In: International Conference on Machine
  Learning (ICML), PMLR, pp 7354--7363

\bibitem[{Zhang et~al(2015)Zhang, Gong, Sch{\"o}lkopf et~al}]{zhang2015multi}
Zhang K, Gong M, Sch{\"o}lkopf B, et~al (2015) Multi-source domain adaptation:
  A causal view. In: AAAI Conference on Artificial Intelligence (AAAI), pp
  3150--3157

\bibitem[{Zhang et~al(2020{\natexlab{b}})Zhang, Jiang, Wang, Kuang, Zhao, Zhu,
  Yu, Yang, and Wu}]{DBLP:conf/mm/ZhangJWKZZYYW20}
Zhang S, Jiang T, Wang T, et~al (2020{\natexlab{b}}) Devlbert: Learning
  deconfounded visio-linguistic representations. In: {MM} '20: The 28th {ACM}
  International Conference on Multimedia. {ACM}, pp 4373--4382

\bibitem[{Zhang et~al(2019{\natexlab{b}})Zhang, Liu, Long, and
  Jordan}]{Zhang2019BridgingTA}
Zhang Y, Liu T, Long M, et~al (2019{\natexlab{b}}) Bridging theory and
  algorithm for domain adaptation. In: International Conference on Machine
  Learning (ICML)

\bibitem[{Zhang et~al(2020{\natexlab{c}})Zhang, Wei, Wu, Zhao, Niu, Huang, and
  Tan}]{zhang2020collaborative}
Zhang Y, Wei Y, Wu Q, et~al (2020{\natexlab{c}}) Collaborative unsupervised
  domain adaptation for medical image diagnosis. IEEE Transactions on Image
  Processing (TIP) 29:7834--7844

\bibitem[{Zhao et~al(2018)Zhao, Zhang, Wu, Moura, Costeira, and
  Gordon}]{zhao2018adversarial}
Zhao H, Zhang S, Wu G, et~al (2018) Adversarial multiple source domain
  adaptation. Advances in Neural Information Processing Systems (NeurIPS)
  31:8559--8570

\bibitem[{Zhao et~al(2020)Zhao, Gong, Liu, Fu, and Tao}]{Zhao2020DomainGV}
Zhao S, Gong M, Liu T, et~al (2020) Domain generalization via entropy
  regularization. In: Advances in Neural Information Processing Systems
  (NeurIPS)

\bibitem[{Zhao et~al(2021)Zhao, Li, Xu, Yue, Ding, and Keutzer}]{zhao2021madan}
Zhao S, Li B, Xu P, et~al (2021) Madan: multi-source adversarial domain
  aggregation network for domain adaptation. International Journal of Computer
  Vision (IJCV) pp 1--26

\bibitem[{Zheng and Yang(2021)}]{zheng2021rectifying}
Zheng Z, Yang Y (2021) Rectifying pseudo label learning via uncertainty
  estimation for domain adaptive semantic segmentation. International Journal
  of Computer Vision (IJCV) 129(4):1106--1120

\bibitem[{Zhou et~al(2020)Zhou, Yang, Hospedales, and Xiang}]{zhou2020learning}
Zhou K, Yang Y, Hospedales T, et~al (2020) Learning to generate novel domains
  for domain generalization. In: European Conference on Computer Vision (ECCV),
  pp 561--578

\bibitem[{Zhou et~al(2021{\natexlab{a}})Zhou, Loy, and Liu}]{zhou2021semi}
Zhou K, Loy CC, Liu Z (2021{\natexlab{a}}) Semi-supervised domain
  generalization with stochastic stylematch. arXiv preprint arXiv:210600592

\bibitem[{Zhou et~al(2021{\natexlab{b}})Zhou, Yang, Qiao, and
  Xiang}]{zhou2021domainadaptive}
Zhou K, Yang Y, Qiao Y, et~al (2021{\natexlab{b}}) Domain adaptive ensemble
  learning. IEEE Transactions on Image Processing 30:8008--8018

\bibitem[{Zhou et~al(2021{\natexlab{c}})Zhou, Yang, Qiao, and
  Xiang}]{zhou2021domain}
Zhou K, Yang Y, Qiao Y, et~al (2021{\natexlab{c}}) Domain generalization with
  mixstyle. In: International Conference on Learning Representations (ICLR)

\bibitem[{Zuo et~al(2021)Zuo, Yao, and Xu}]{zuo2021attention}
Zuo Y, Yao H, Xu C (2021) Attention-based multi-source domain adaptation. IEEE
  Transactions on Image Processing 30:3793--3803

\end{thebibliography}


\end{document}